\documentclass[manuscript]{acmart}
\AtBeginDocument{%
  }

\usepackage{booktabs}
\usepackage{graphicx} 
\usepackage{subfigure}
\usepackage[export]{adjustbox}
\begin{document}
\title{Panoptic Perception for Autonomous Driving: A Survey}

\author{Yunge Li}
\email{yungeli@oakland.edu}
\affiliation{%
  \institution{Oakland University}
  \city{Rochester Hills}
  \country{USA}
  \postcode{48309}
}

\author{Lanyu Xu}
\email{lxu@oakland.edu}
\affiliation{%
  \institution{Oakland University}
  \city{Rochester Hills}
  \country{USA}
  \postcode{48309}
  }


\begin{abstract}
Panoptic perception represents a forefront advancement in autonomous driving technology, unifying multiple perception tasks into a singular, cohesive framework to facilitate a thorough understanding of the vehicle's surroundings. This survey reviews typical panoptic perception models for their unique inputs and architectures and compares them to performance, responsiveness, and resource utilization. It also delves into the prevailing challenges faced in panoptic perception and explores potential trajectories for future research. Our goal is to furnish researchers in autonomous driving with a detailed synopsis of panoptic perception, positioning this survey as a pivotal reference in the ever-evolving landscape of autonomous driving technologies.
\end{abstract}

\ccsdesc[300]{Computing methodologies~Object detection}
\ccsdesc[300]{Computing methodologies~Image segmentation}
\ccsdesc[500]{Computing methodologies~Multi-task learning}

\keywords{autonomous driving, panoptic perception}


\maketitle

\section{Introduction}

\subsection{Motivation for panoptic perception in autonomous driving}
In autonomous driving, accurately perceiving and interpreting complex, dynamic environments in real-time is paramount. Traditional methodologies in vehicular perception typically compartmentalize tasks such as object detection, instance segmentation, and semantic segmentation, addressing each in isolation. While these modular approaches yield valuable insights, they fall short of providing an integrated, holistic understanding of the multifaceted driving environment. This limitation underscores the necessity for panoptic perception, an approach designed to unify disparate perception tasks within a comprehensive framework. The concept of panoptic perception was first introduced in the YOLOP\cite{wu2022yolop} in 2021, marking a significant advancement in autonomous driving. Before this, and continuing presently, numerous multi-task networks have been developed with similar objectives, striving to enhance environmental perception in autonomous driving by unifying various perception tasks within a single, cohesive framework. Panoptic perception, in essence, aligns closely with the principles of multi-task models, sharing many foundational aspects. "Panoptic perception network" and "multi-task network" are used interchangeably in this survey due to their conceptual and functional similarities.

Panoptic segmentation\cite{kirillov2019panoptic}, an amalgamation of instance and semantic segmentation, exemplifies this integrated approach. It has gained widespread application in various visual tasks for its ability to more comprehensively and distinctly identify "things" and "stuff" within a scene, such as roads, sky, buildings, vehicles, pedestrians, and traffic signs. Panoptic perception affords a more detailed and comprehensive environmental view by cohesively combining tasks like object detection, lane line segmentation, drivable area segmentation, semantic segmentation, instance segmentation, and depth estimation. This enriched perception plays a crucial role in enhancing the decision-making capabilities of autonomous driving systems, enabling them to respond more effectively to the intricacies of real-world driving scenarios.

Several key motivations drive the integration of panoptic perception into autonomous driving systems, each contributing to a more reliable and efficient vehicular perception system. Firstly, panoptic perception significantly enhances the robustness and accuracy of environmental understanding. This improvement stems from the ability of the multi-task network in panoptic systems to establish interconnections between individual perception tasks, effectively compensating for their isolated limitations. Furthermore, implementing multi-modal, multi-task networks augments the perception system’s output by fusing multiple input modalities, leading to a more robust and precise sensory interpretation.

The second motivation lies in the optimization of processing efficiency and the reduction of computational overhead. Employing a shared backbone across different tasks within the multi-task network minimizes redundant computations for identical objects and regions. Shared backbone not only streamlines the feature extraction process but also significantly improves the overall efficiency of the model.

Panoptic perception enhances the autonomous vehicle's intelligent, context-aware decision-making capacity. Perception serves as a critical precursor to the decision-making phase in autonomous driving, with the fidelity of perception data directly influencing the vehicle's motion choices and actions. By providing a comprehensive and detailed perception of the environment, panoptic perception equips the vehicle with the necessary insights to make informed and situationally appropriate decisions.

\subsection{Objective and Organization}
Presently, various traditional methodologies have been successfully implemented in autonomous driving, particularly within the scope of Level 3 automation. These conventional approaches typically rely on single, modular models. However, panoptic perception emerged as an innovative and increasingly significant technology in this field. It distinguishes itself through superior robustness, enhanced accuracy, and heightened efficiency compared to traditional methods. Despite its growing relevance, there is a notable absence in the literature of a comprehensive collection and analysis of knowledge of panoptic perception, specifically within the autonomous driving context. This survey endeavors to bridge this gap.

Our objective is to elucidate the fundamental concepts and theories underpinning panoptic perception and to offer an in-depth analysis of the currently prevalent panoptic perception models. By doing so, we aim to provide a more defined framework and direction for researchers delving into this area. We anticipate that this survey will catalyze further studies and discussions, contributing to the evolution of panoptic perception research in autonomous driving.

Literature search and collection were conducted on \textbf{Google Scholar}, \textbf{Scopus}, and \textbf{Arxiv}. We used the following keywords to search: \textbf{autonomous driving}, \textbf{multi-task learning}, \textbf{panoptic perception}, and collected approximate 218 papers in total. In addition, papers published before June 2023 were considered for inclusion in this survey. Those papers need to be more detailed. Papers cannot be cited in full and need to be filtered based on whether they answer the following questions:\\
1) What tasks are mainly included in panoptic perception networks?\\
2) Does the model outperform the single-task performance baseline in some way? \\
3) Whether to compare with other panoptic perception models and has certain outstanding points. \\
\vspace{-2mm}
\begin{figure}[htb]
  \centering
  \includegraphics[scale=0.4]{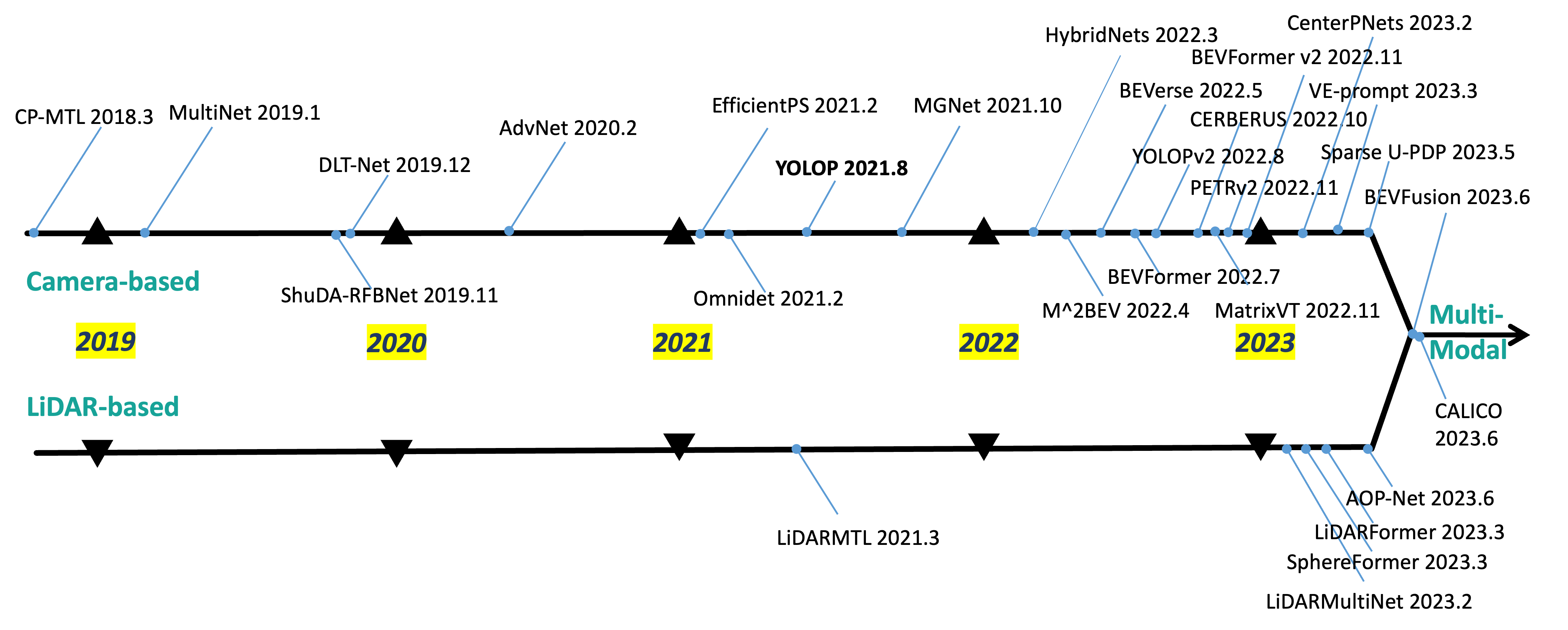}
  \caption{Overview of multi-task perception model for autonomous driving}
  \label{fig:model history}
\end{figure}
\vspace{-4mm}
Within the scope of this survey, 28 out of the 218 reviewed papers presented unique panoptic perception networks, addressing the aforementioned questions. This survey meticulously collates these 28 publications, providing a detailed analysis and comparative study of the various networks they introduce. Figure~\ref{fig:model history} contains a complete list of all models in the order of their publication time. These models, pivotal in the realm of panoptic perception, primarily differ in their input sources and leverage a range of architectural frameworks, including convolutional neural networks (CNNs), transformers, and hybrid models, to yield promising results in their respective applications. Referring to the taxonomy of the survey of 3D object detection \cite{qian20223d}, existing efforts can be divided into the following three subdivisions: 1) image-based such as \cite{wu2022yolop}, \cite{han2022yolopv2}, \cite{vu2022hybridnets}, \cite{qian2019dlt}, \cite{teichmann2018multinet}, \cite{mohan2021efficientps}, \cite{chen2023centerpnets}, \cite{schon2021mgnet}, \cite{wang2019shuda}, \cite{liu2019advnet}, \cite{scribano2022cerberus}, \cite{wang2023sparse}, \cite{xie2022m}, \cite{kumar2021omnidet}, \cite{chen2018multi}, \cite{philion2020lift}, which are currently more commonly used methods in the field of panoptic perception, because images can provide more rich features; 2) point cloud based such as \cite{feng2021simple}, \cite{xu2023aop}, \cite{zhou2023lidarformer}, \cite{ye2023lidarmultinet}, \cite{lai2023spherical}, these methods make up for the disadvantage of not obtaining accurate depth information; 3) multi-modal fusion \cite{liu2022bevfusion}, \cite{sun2023calico}, which are a current development trend, because this kind of methods obtain accurate depth information while obtaining rich image features, but how to better fuse inputs from different modalities is a challenge.

The remainder of this survey is organized as follows. In Section \ref{Background}, we lay the background by delving into essential aspects of sensors, perception tasks, datasets and benchmark, and evaluation metrics pertinent to panoptic perception. Section \ref{Techniques for Panoptic Perception} reviews approaches to panoptic perception in autonomous driving and their corresponding architectures in detail. Section \ref{Comparison With Other Perception Techniques} provides a comprehensive comparison of the state-of-the-art perceptual networks. We discuss the current challenges in panoptic perception and identify future research directions in Section \ref{Challenges and Future Directions}. Finally, we conclude this paper in Section \ref{Conclusion}.

\section{Background} \label{Background}
\subsection{Hardware for Panoptic Perception}

Autonomous driving heavily relies on the ability of a vehicle to perceive and understand its environment. This is achieved through the use of various sensors that capture data about the surroundings of the vehicle. In this section, we will explore the different types of sensors that are commonly used in autonomous driving perception and their respective strengths and weaknesses. Moreover, we also study the existing sensor fusion methods to overcome the errors caused by the shortcomings of single-type sensors.
\begin{table}[!htb]
\resizebox{\textwidth}{!}{
\begin{tabular}{lllll}
\hline
\multicolumn{1}{c}{Sensors} &
   &
  \multicolumn{1}{c}{Advantages} &
  \multicolumn{1}{c}{Disadvantages} \\ \hline
 &
  \multicolumn{1}{c}{Monocular Camera} &
  \begin{tabular}[c]{@{}l@{}}$\bullet$ Easy to install and low cost.\\ $\bullet$ Get rich color and texture information.\\ \\ \\ \end{tabular} &
  \begin{tabular}[c]{@{}l@{}}$\bullet$ Depth cannot be measured directly.\\ $\bullet$ 2D and 3D perception is limited.\\ $\bullet$ Affected by lighting and environmental changes.\\ \\ \end{tabular} &
   \\ \cline{2-5} 
Camera &
  Stereo Camera &
  \begin{tabular}[c]{@{}l@{}}$\bullet$ Estimate depth information.\\ $\bullet$ Get rich color and texture information.\\ \\ \\ \end{tabular} &
  \begin{tabular}[c]{@{}l@{}}$\bullet$ Complex algorithms are required.\\ $\bullet$ Consume more computing power.\\ $\bullet$ Affected by lighting and environmental changes.\\ \\ \end{tabular} &
  \\ \cline{2-5} 
 &
  Fish-eye Camera &
  \begin{tabular}[c]{@{}l@{}}$\bullet$ Provide a wider field of view.\\ $\bullet$ Get rich color and texture information.\\ \\ \\ \end{tabular} &
  \begin{tabular}[c]{@{}l@{}}$\bullet$ There is severe distortion in the image.\\ $\bullet$ Require complex correction algorithms. \\ $\bullet$ Affected by lighting and environmental changes.\\ \\ \end{tabular} &
  \\ \hline
 &
  1D LiDAR &
  \begin{tabular}[c]{@{}l@{}}$\bullet$ Measure depth information directly.\\ $\bullet$ Robust to illumination and environmental changes.\\ \\ \end{tabular} &
  \begin{tabular}[c]{@{}l@{}}$\bullet$ Only one-dimensional information can be obtained.\\ $\bullet$ Limited color and texture information.\\ \\ \end{tabular} &
  \\ \cline{2-5} 
LiDAR &
  2D LiDAR &
  \begin{tabular}[c]{@{}l@{}}$\bullet$ Measure depth information directly.\\ $\bullet$ Robust to illumination and environmental changes.\\ $\bullet$ The two-dimensional information can be obtained.\\ \\ \end{tabular} &
  \begin{tabular}[c]{@{}l@{}}$\bullet$ Vertical information cannot be obtained.\\ $\bullet$ 3D perception of complex environments is limited.\\ $\bullet$ Limited color and texture information.\\ \\ \end{tabular} &
  \\ \cline{2-5} 
 &
  3D LiDAR &
  \begin{tabular}[c]{@{}l@{}}$\bullet$ Measure depth information directly.\\ $\bullet$ Robust to illumination and environmental changes.\\ $\bullet$ Three-dimensional information can be obtained.\\ \\ \end{tabular} &
  \begin{tabular}[c]{@{}l@{}}$\bullet$ The cost is high.\\ $\bullet$ Data processing is complicated. \\ $\bullet$ Limited color and texture information.\\ \\ \end{tabular} &
 \\ \hline
\end{tabular}}
\caption{Advantages and disadvantages of different sensors}
\label{table:sensors}
\end{table}
\vspace{-25pt}
\subsubsection{Cameras}
\ 
\newline
Cameras are one of the most common sensors used in autonomous driving. They capture visual data about the environment in the form of images. Cameras are versatile and can be used in various applications such as object detection, traffic sign recognition, and lane segmentation. They can also provide contextual information about the environment, such as lighting conditions, weather, and road conditions.
\vspace{-10pt}
\begin{figure}[htbp]
  \centering
      \subfigure[Basler acA1600-60gc~\cite{acA1600-60gc}] {\includegraphics[width=.18\textwidth]{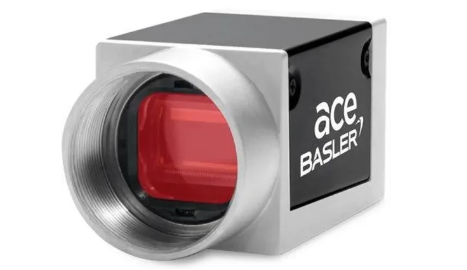}\label{mono camera}}
      \hspace{10mm}
      \subfigure[ZED 2 Stereo~\cite{stereolabs_zed2}]{\includegraphics[width=.2\textwidth]{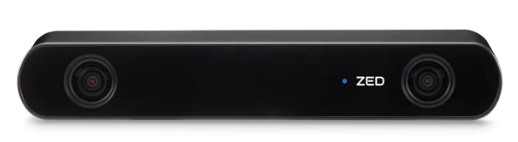}\label{stereo camera}}
      \hspace{10mm}
      \subfigure[Wide-angle~\cite{widecamera}]{\includegraphics[width=.12\textwidth]{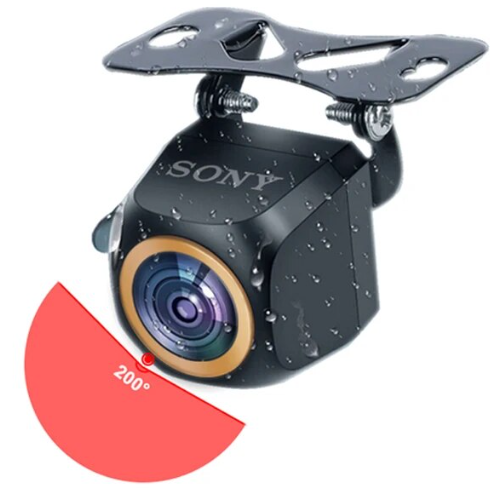}\label{wide camera}}
  \vspace{-10pt}
  \caption{Cameras}
  \label{fig:mono camera}
\end{figure}
\vspace{-10pt}

\textbf{Monocular Cameras}. Monocular cameras use a single lens to capture images of the environment. They are the simplest and most widely used camera types in autonomous driving. Monocular cameras are typically small and lightweight, making them ideal for use in small vehicles or drones. They are also relatively cheap and have low power requirements. However, they have limited depth perception and cannot accurately estimate the distance of objects in the environment. The most commonly used monocular camera in autonomous driving is the Basler acA1600-60gc~\cite{acA1600-60gc}, which has a high resolution of 1.92 megapixels and is suitable for applications that require precise images. In addition, Basler acA4112-30uc, which has a higher resolution of 12 megapixels, can capture extremely fine details. However, it may have high requirements for processing time and computing power.

\textbf{Stereo Cameras}.
Stereo cameras use two lenses to capture images of the environment. They are used to create stereoscopic images that provide depth perception. This allows them to estimate the distance of objects in the environment accurately. Stereo cameras are larger and heavier than monocular cameras, so they have certain requirements on the space and load of the vehicle. In addition, environmental information analysis and depth estimation of stereo cameras also require some complex algorithms and more powerful computing power, so they are more expensive and have higher power requirements. The commonly used Bumblebee2 BB2-08S2C is easy to use, stable, and produces high-quality stereoscopic images but low resolution. The ZED Stereo Camera~\cite{stereolabs_zed2} can provide accurate depth images and three-dimensional views. However, performance may be limited in low-light environments because it requires sufficient lighting. 

\textbf{Wide-angle or Fish-eye Cameras}. 
Wide-angle or fisheye cameras capture a wide field of view and are ideal for use in areas that require a wide view of the environment. They can capture images up to 200 degrees, which is a much wider field of view than monocular or stereo cameras. It can also obtain depth information through specific algorithms. However, they have significant distortion, making it difficult to estimate the distance of objects in the environment accurately, so complex algorithms need to be applied to account for such errors. Wide-angle or fisheye cameras are often used in low-speed applications such as parking or maneuvering in confined spaces. In the application, A SONY FishEye Camera~\cite{widecamera} is shown in the figure~\ref{wide camera} has a large field of view of 200 degrees, which can provide a full range of vision.
\subsubsection{LiDAR}
\ 
\newline
LiDAR is a remote sensing technology that uses lasers to create a 3D environment representation. It measures the time it takes for a laser pulse to reflect off an object and return to the sensor, allowing it to estimate the distance of objects in the environment accurately. LiDAR provides highly accurate 3D environmental information, making it ideal for depth estimation and other 3D-related tasks. One advantage of LiDAR over cameras is its ability to operate in low light conditions, making it a preferred sensor for driving at night. However, LiDAR can be affected by adverse weather conditions such as rain or fog, which can scatter the laser beam and reduce its accuracy.
\vspace{-10pt}
\begin{figure}[htbp]
  \centering
  \subfigure[Garmin 1D LiDAR-Lite v3~\cite{garmin_product}] {\includegraphics[width=.10\textwidth]{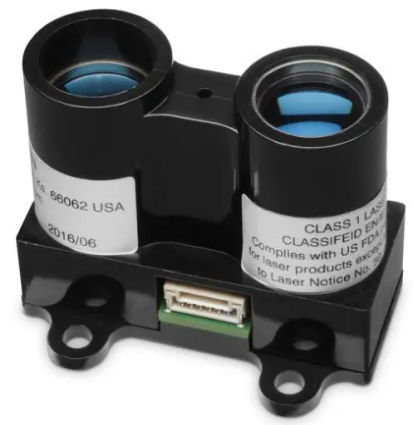}\label{1D LiDAR Cameras}}
  \hspace{15mm}
  \subfigure[2D LiDAR Sick LMS1xx~\cite{sick_lidar_sensor}]{\includegraphics[width=.16\textwidth]{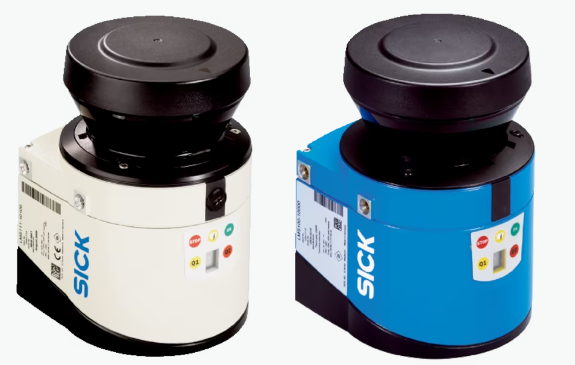}\label{2D LiDAR Cameras}}
  \hspace{15mm}
  \subfigure[3D LiDAR Ouster OS1-64~\cite{3D_Lidar_ouster}]{\includegraphics[width=.12\textwidth]{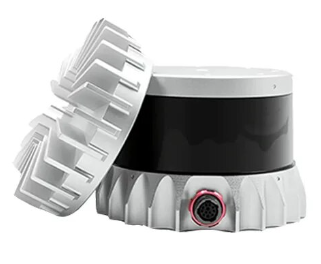}\label{3D LiDAR Cameras}}
  \vspace{-10pt}
  \caption{LiDARs}
  \label{fig:mono camera}
\end{figure}
\vspace{-10pt}

\textbf{1D LiDAR}.
1D LiDAR uses a single laser beam to measure the distance of objects in one direction. It is typically used in low-speed applications, such as collision avoidance systems. 1D LiDAR provides accurate distance measurements but has limited coverage and cannot provide a comprehensive 3D representation of the environment. In practice, the Garmin LiDAR-Lite v3~\cite{garmin_product} is a lightweight and affordable 1D LiDAR for drones, robotics, and other automation scenarios.

\textbf{2D LiDAR}.
2D LiDAR uses rotating laser light to measure the distance of objects in a two-dimensional plane. This feature enables it to create a 2D representation of the environment. 2D LiDAR is commonly used in autonomous driving for obstacle detection and mapping. For example, Sick LMS1~\cite{sick_lidar_sensor} is a 2D LiDAR sensor with very high reliability and precision, suitable for outdoor environments, and has certain environmental adaptability. 2D LiDAR provides a more comprehensive view of the environment than 1D LiDAR but with limited vertical coverage.

\textbf{3D LiDAR}.
3D LiDAR uses multiple lasers to measure the distance of objects in a three-dimensional space. This allows it to create a highly accurate 3D representation of the environment. 3D LiDAR is commonly used in autonomous driving and provides the most accurate and comprehensive representation of the environment. However, 3D LiDAR is more expensive and requires higher power than 1D or 2D LiDAR. In practical applications, Waymo's proprietary LiDAR sensors can provide high-resolution environmental perception data, especially in autonomous driving applications. However, such proprietary equipment can be relatively expensive and not readily available. Ouster OS1-64~\cite{3D_Lidar_ouster} is another high-performance 64-channel LiDAR, which can provide rich 3D information and a more comprehensive understanding of the environment, but the price is still high.
\subsubsection{Sensor Fusion} \label{sensorfusion}
\ 
\newline
Cameras and LiDAR are two of the most important sensors used in autonomous driving. Each type of sensor has its own strengths and weaknesses, which must be considered when designing an autonomous driving system. By understanding the different types of sensors, we can design more effective autonomous driving systems that can perceive and understand the environment with greater accuracy and reliability. Sensor fusion is a common way to overcome the shortcomings of a single sensor. The fusion of multiple sensor data can obtain more accurate input data to improve the performance of various perception tasks. Fusion strategies are usually divided into early fusion, mid-term fusion, and late fusion, which perform fusion in the data input, feature extraction, and decision-making phases, respectively. 

\begin{figure}[htbp]
  \centering
  \subfigure[Early fusion (Data fusion)] {\includegraphics[width=.465\textwidth, frame]{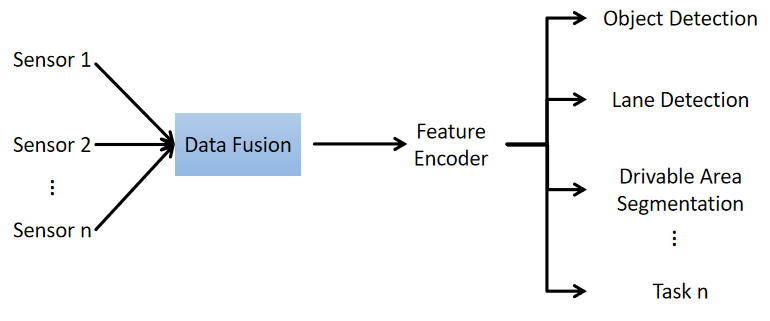}\label{fig:data fusion}}
  \subfigure[Mid-term fusion (Feature fusion)]{\includegraphics[width=.49\textwidth, frame]{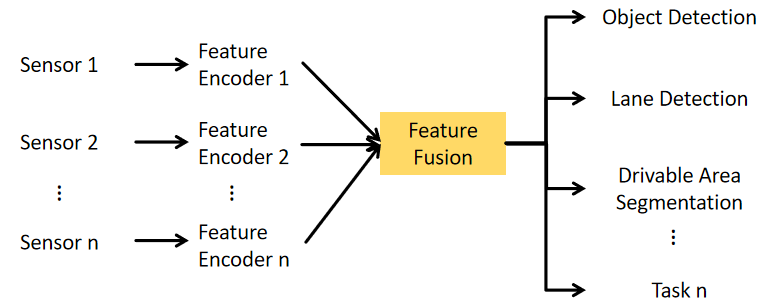}\label{fig:feature fusion}}
  \subfigure[Late fusion (Decision-level fusion)]{\includegraphics[width=.6\textwidth, frame]{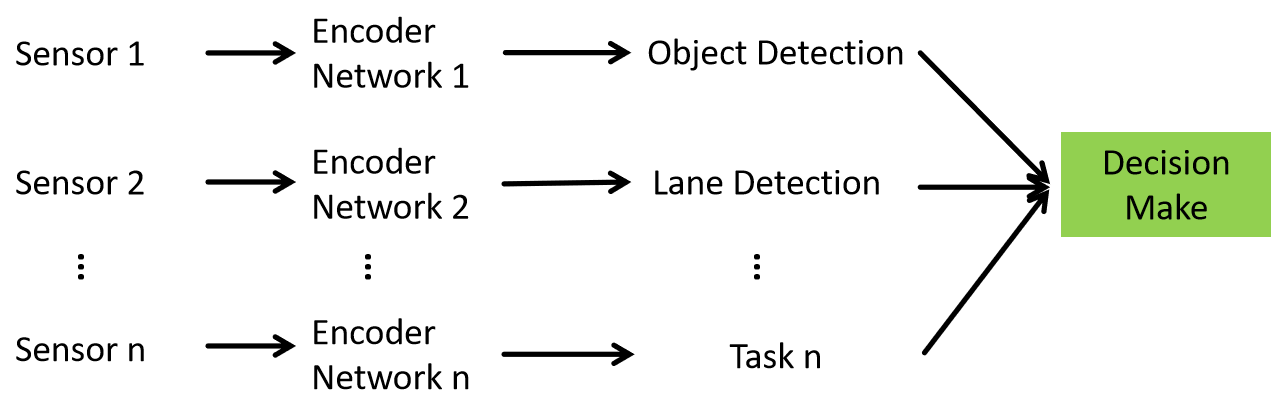}\label{fig:late fusion}}
  \vspace{-10pt}
  \caption{Sensor fusion}
  \label{Sensor fusion}
\end{figure}

\textbf{Early fusion} needs to deal with data format and scale issues. As shown in figure~\ref{fig:data fusion}, the data from different sensors are first fused, and then the fused data is put into the feature encoder for feature extraction and then the corresponding tasks. 
\vspace{-10pt}
\textbf{Mid-term fusion} in sensor data processing involves designing a network architecture capable of integrating features from various sources. As illustrated in Figure~\ref{fig:feature fusion}, data from different sensors undergo separate feature extraction processes. For instance, camera data (images) are typically processed by neural networks like ResNet or VGG for feature encoding, whereas LiDAR data are handled by networks such as VoxelNet\cite{zhou2018voxelnet}, specifically designed for point cloud encoding. A significant challenge in mid-term fusion is reconciling the different feature dimensions from camera and LiDAR data. To overcome this, a prevalent approach involves converting both types of features into Bird's Eye View (BEV) feature maps. This transformation allows for the fusion of these diverse data types in a unified dimensional space. The BEV feature maps provide a more comprehensive and richer environmental representation, enhancing the system’s ability to perform specific tasks with greater accuracy and efficiency. \textbf{Late fusion} focuses on integrating these different data types at the decision-making stage of the process. This approach allows each sensor type to independently process and interpret data before combining their results for final decision-making. Late fusion is rarely applied to networks at the perception stage that does not involve making decisions.

Given the pros and cons of camera and LiDAR, there are an increasing number of works in panoptic perception that adopt sensor fusion strategies to seek better performance. In the realm of panoptic perception, mid-term fusion is particularly common. It effectively merges the detailed image features captured by cameras with the spatially rich point cloud features from LiDAR, creating a more complete and accurate representation of the environment. More detail will be introduced in Section\ref{sensor fusion network}.
\subsection{Panoptic Perception Tasks} \label{Panoptic Perception Tasks}

This survey mainly focuses on six different tasks, including object detection (OD), lane segmentation (LD/LS), drivable area segmentation (DAS), instance segmentation (IS), semantic segmentation (SS), and depth estimation (DE), because these tasks are the most common in environmental understanding and perception.  \\
\textbf{Object Detection:}
Object detection aims to find all objects in an image and determine their category and location. This task is usually divided into the following four types of solutions according to the stages and anchors' strategies. \\
\emph{One-stage, Anchor-based}: This approach completes object classification and localization in one step, using predefined anchor boxes to generate bounding boxes. A typical example is You Only Look Once (YOLO)\cite{redmon2016you}. YOLO divides the input image into many same-dimensional grids and predicts each grid cell's bounding box and category. Single Shot MultiBox Detector (SSD)\cite{liu2016ssd} is also this type of method, which generates bounding boxes at multiple scales to detect objects of different sizes. Such methods have the advantage of being computationally efficient but may not be as accurate as using a two-stage approach.\\
\vspace{-5pt}
\begin{figure}[htb]
  \centering
  \includegraphics[scale=0.4]{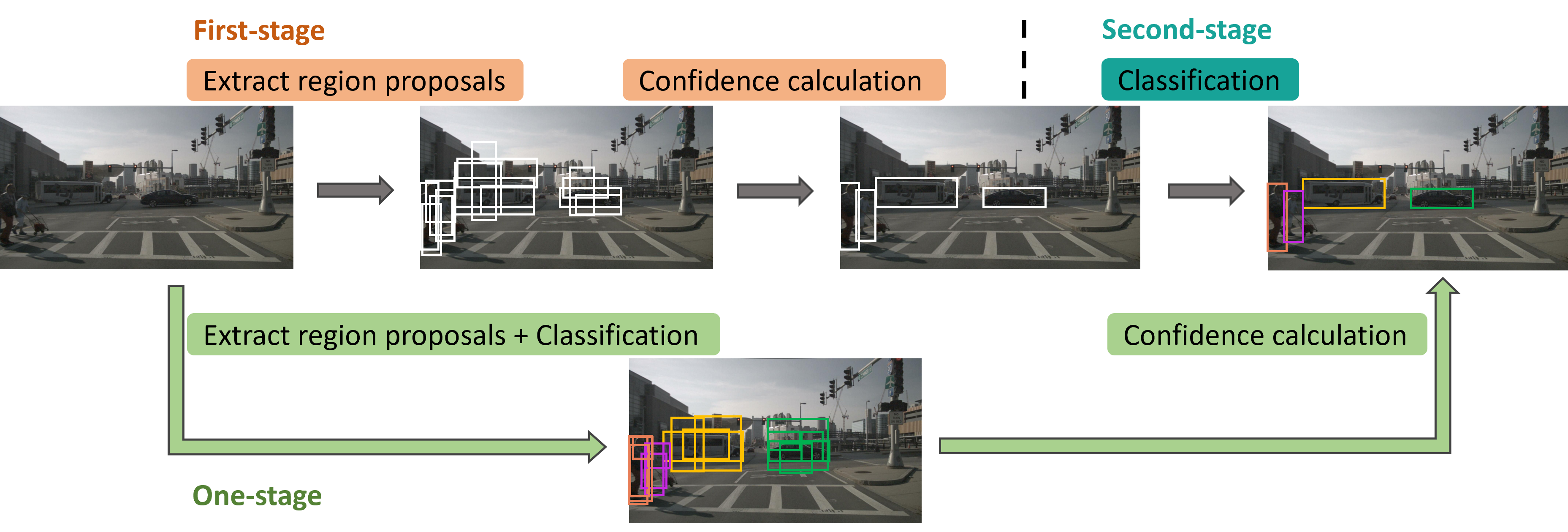}
  \caption{Overview of anchor-based object detection models.}
  \label{fig:anchorbased detection}
\end{figure}
\vspace{-10pt}
\emph{Two-stage, Anchor-based}: In this type of method, the first stage generates region proposals that may contain objects. In the second stage, the regions are classified, and the bounding box is corrected. A well-known example is R-CNN\cite{girshick2014rich}, which utilizes the Region Proposal Network (RPN) to generate region proposals and extract features for classification and regression. These techniques have the benefit of offering high accuracy, and they can handle more complex objects and backgrounds. However, they are less computationally efficient.\\
\emph{One-stage, Anchor-free}: This method also completes object classification and localization in one step but does not use predefined anchor boxes. A typical example is CenterNet\cite{duan2019centernet}, which uses keypoint detection to locate the center point of an object and predict the size and offset of the object. CenterPoint\cite{yin2021center} is also a typical application in 3D object detection. Such methods have the advantage of not needing to set anchor points and are more flexible to handle objects of various shapes and sizes but may require more complex post-processing steps.\\
\emph{Two-stage, Anchor-free}: These methods generate region proposals that may contain objects in the first stage but do not use predefined anchor boxes. Corner Proposal Network (CPN)\cite{duan2020corner} is a network that uses a corner heatmap to complete object detection. It first predicts corners to compose several object proposals and then applies second-stage classification to filter out false positives and assign a class label for each survived proposal.\\
\emph{Transformer-based}: In addition, Detection Transformer (DETR) \cite{carion2020endtoend} represents a different paradigm in the field of object detection, which gets rid of traditional anchor boxes and post-processing steps, such as non-maximum suppression (NMS). DETR treats the object detection problem as a set prediction problem. Each output of the decoder corresponds to a potential object, which is either an actual object category or a "no object" category, used to indicate that no object was detected. 
\vspace{-5pt}
\begin{figure}[htb]
  \centering
  \includegraphics[scale=0.4]{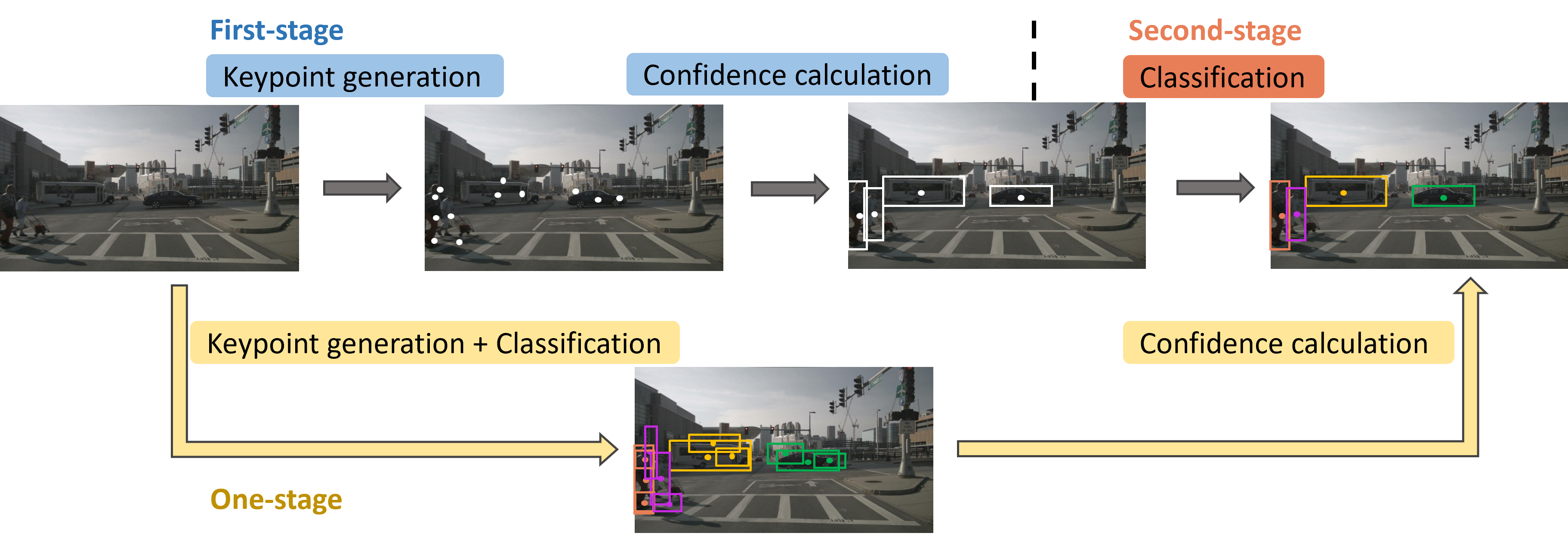}
  \caption{Overview of anchor-free object detection models.}
  \label{fig:anchorfree detection}
\end{figure}
\vspace{-10pt}

\emph{Loss function}: These CNN-based object detection methods usually use cross-entropy loss for classification and regression loss, such as Smooth L1 or IoU loss, to optimize bounding boxes. DETR combines the Hungarian algorithm (an optimal allocation algorithm) to match one-to-one relationships between predictions and ground-truth targets and calculates losses for classification and bounding box regression.

\vspace{-0.35cm}
\begin{figure}[htbp]
  \centering
  \subfigure[Instance segmentation] {\includegraphics[width=.22\textwidth]{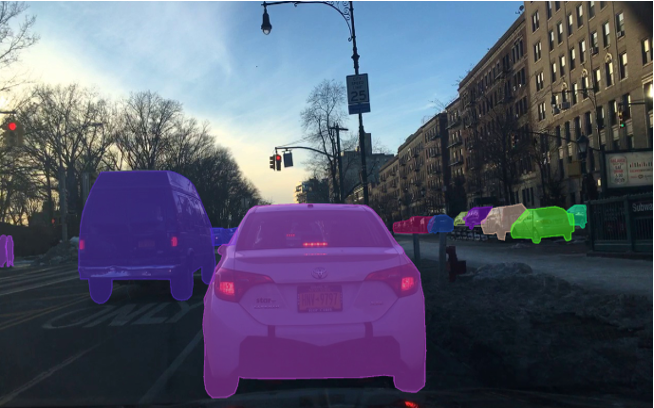}\label{fig:Instance Segmentation}}
  \hspace{.5mm}
  \subfigure[Semantic segmentation]{\includegraphics[width=.22\textwidth]{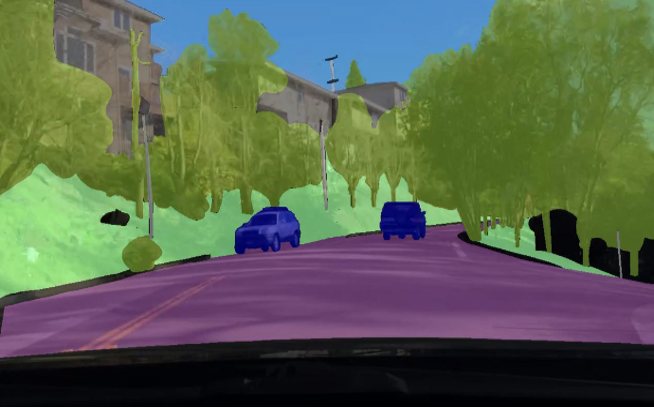}\label{fig:Semantic Segmentation}}
  \hspace{.5mm}
  \subfigure[Lane segmentation]{\includegraphics[width=.22\textwidth]{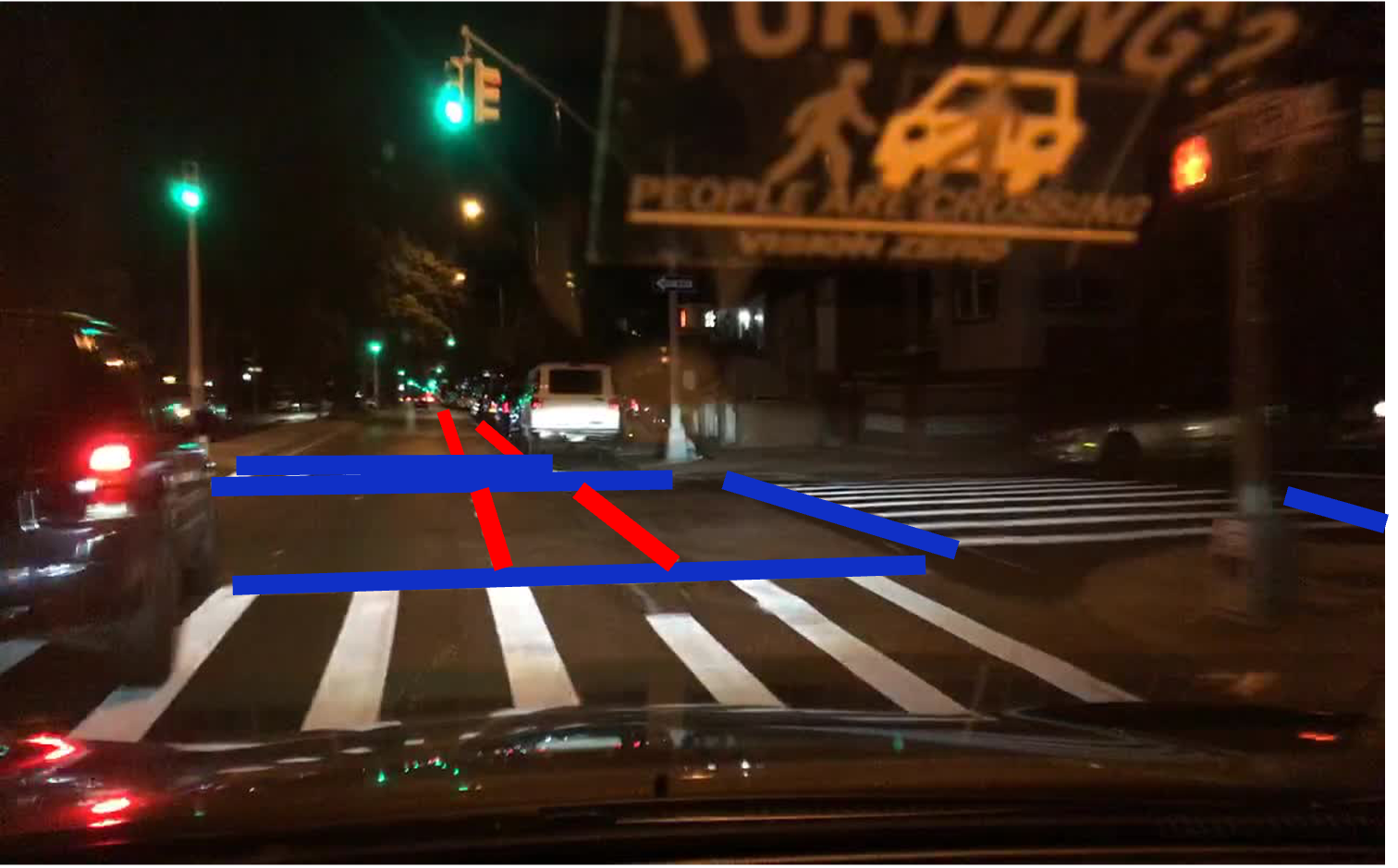}\label{fig:Lane Segmentation}}
  \hspace{.5mm}
  \subfigure[Drivable area Segmentation]{\includegraphics[width=.22\textwidth]{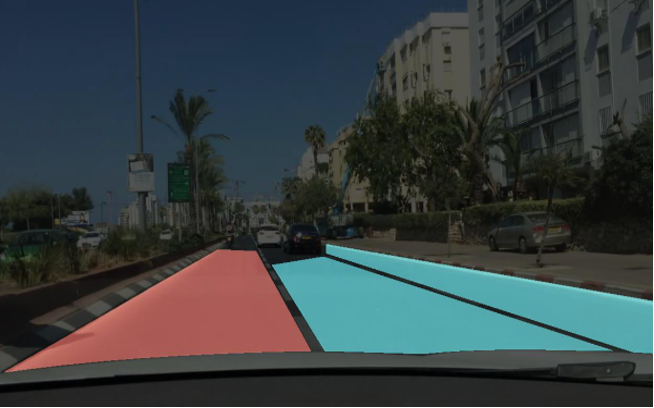}\label{fig:Drivable Area Segmentation}}
  \vspace{-0.2cm}
  \caption{Segmentation tasks~\cite{yu2020bdd100k}}
  \label{Segmentation tasks}
\end{figure}
\vspace{-10pt}

\textbf{Segmentation:}
\emph{Instance segmentation} and \emph{semantic segmentation} play crucial roles in the perception systems of autonomous vehicles, aiding in the precise understanding and analysis of complex environments. Instance segmentation identifies each independent object within an image and accurately separates it from the background, such as pedestrians and vehicles, while semantic segmentation classifies each pixel into different semantic categories, such as road surfaces, pedestrians, and vehicles, providing detailed information for safe driving.
The main challenges faced by these segmentation tasks include class imbalance and the difficulty of segmenting small objects, which can impact the model's accuracy and efficiency. To address these issues, various loss functions, such as cross-entropy loss, Focal Loss~\cite{lin2017focal}, Dice Loss~\cite{sudre2017generalised}, and Tversky Loss~\cite{salehi2017tversky}, have been employed to optimize classification accuracy and enhance the recognition of small objects. Notably, they effectively resolve class imbalance by focusing the model's learning on more challenging and less frequent cases.
\emph{Lane segmentation} and \emph{drivable area segmentation} are specific segmentation tasks in autonomous driving, requiring the model to differentiate between lanes and drivable areas, which is crucial for ensuring the correct driving path of the vehicle. These tasks pose challenges in complex traffic or varying lighting conditions, necessitating exact segmentation capabilities. Researchers have explored optimizing Fully Convolutional Networks (FCN) by using multiple loss functions~\cite{wu2022yolop},~\cite{han2022yolopv2},~\cite{vu2022hybridnets} and multi-modal information fusion techniques to enhance task performance, combining the advantages of image and LiDAR data for a richer environmental understanding.
In summary, despite progress in segmentation applications in autonomous driving, future research will focus on optimizing models to handle class imbalance, improve the accuracy of small object segmentation, and enhance lane and drivable area recognition capabilities. By refining loss functions and integrating various perception information, the performance of autonomous vehicle perception systems can be further improved, ensuring their safety and reliability in diverse and unpredictable driving environments.

\textbf{Depth Estimation:}
Depth Estimation, a fundamental task in computer vision, aims to accurately predict the depth value of each pixel in an image. This task is commonly approached in two ways: through depth map regression or by classifying each pixel into discrete depth levels, particularly in systems that utilize multi-view images as input. These techniques are pivotal in creating a 3D scene representation from 2D images.
\begin{figure}[htb]
  \centering
  \includegraphics[scale=0.4]{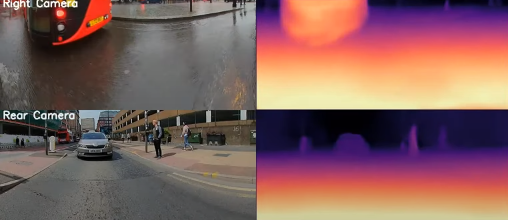}
  \caption{Depth estimation example for Woodscape\cite{yogamani2019woodscape}.}
  \label{fig:Depth Estimation}
\end{figure}
\vspace{-10pt}

In contrast, when using LiDAR data as input, the process primarily involves handling and interpreting point cloud data. LiDAR provides direct 3D spatial information, simplifying the task of depth estimation compared to methods relying solely on image data.
Regardless of the input type, the loss function in a depth estimation network is crucial. It typically measures the discrepancy between the predicted depth values and the true depth values. Common loss functions used in this context include Mean Square Error (MSE) Loss and Huber Loss~\cite{Huberloss}. MSE Loss is widely used for its simplicity and effectiveness in penalizing large errors. On the other hand, Huber Loss combines the properties of both MSE and Mean Absolute Error (MAE), offering a more robust approach against outliers, which is particularly beneficial in environments with varying depth ranges or in the presence of noise in the data.
\subsection{Benchmark and Dataset}
In panoptic perception networks for autonomous driving, the importance of datasets is self-evident because they provide the necessary samples for model training and validation and are the cornerstone of multi-task learning while also allowing models to perform generalization evaluation and benchmark tests. They provide researchers with the convenience of comparing different algorithms and models under uniform conditions. In addition, the high-quality dataset simulates complex real-world scenes, allowing the model to be effectively trained in changing weather, lighting, road environment, object behavior, etc. Therefore, selecting and using an appropriate dataset profoundly and critically impacts model performance. In this section, we introduce the most popular datasets regarding size, diversity, annotation, limitation, and their supporting tasks. Table~\ref{table:dataset} and ~\ref{dataset tasks} visually demonstrate the content of this section. \\

\textbf{Data Types}. Autonomous driving datasets can have various types of inputs. Monocular camera images are the most common type, and datasets like COCO\cite{lin2014microsoft}, ADE20K\cite{zhou2017scene}, Mapillary Vistas\cite{neuhold2017mapillary}, Indian Driving Dataset (IDD)\cite{varma2019idd}, Cityscapes\cite{cordts2016cityscapes}, BDD100K\cite{yu2020bdd100k}, and KITTI\cite{geiger2012we} are popular for this. Stereo camera datasets are relatively fewer, but KITTI and ApolloScape\cite{huang2018apolloscape} provide binocular vision data. Fisheye cameras can capture a wider perspective, but datasets for this camera are limited. WoodScape\cite{yogamani2019woodscape} is the first extensive fisheye automotive dataset. KITTI and ApolloScape are popular datasets for 3D point cloud data. Large-scale autonomous driving datasets such as Waymo Open Dataset\cite{sun2020scalability}, nuScenes\cite{caesar2020nuscenes}, and Lyft Level 5\cite{houston2021one} include 3D LiDAR data and camera images and are widely used.
\vspace{-5pt}
\begin{table}[!htb]
\resizebox{\textwidth}{!}{
\begin{tabular}{lllllllllcllcclclllclll}
\cline{1-17}
Dataset &
  Year &
  \multicolumn{1}{l}{Size} &
   &
   &
  Diversity &
  \multicolumn{1}{l}{} &
   &
   &
  \multicolumn{1}{l}{} &
  \multicolumn{1}{l}{} &
   &
  \multicolumn{1}{l}{Modality} &
   &
   &
   &
  \multicolumn{1}{l}{Accessibility} &
   &
   &
   \\ \cline{3-4} \cline{6-11} \cline{13-16}
 &
   &
  \multicolumn{1}{l}{Image} &
  \multicolumn{1}{l}{Video} &
   &
  Classes &
  \multicolumn{1}{l}{Sunny} &
  Rainy &
  Snowy &
  \multicolumn{1}{l}{Daily} &
  \multicolumn{1}{l}{Night} &
   &
  \multicolumn{1}{l}{Monocular} &
  Stereo &
  Fish-eye &
  3D LiDAR &
  \multicolumn{1}{l}{} &
   &
   &
   \\ \cline{1-17}
COCO\cite{lin2014microsoft} &
  2014 &
  200k &
  - &
   &
  171 &
  \checkmark &
   &
   &
  YES &
  NO &
   &
  \checkmark &
   &
   &
   &
  Free &
   &
   &
   \\
ADE20K\cite{zhou2017scene} &
  2017 &
  25k &
  - &
   &
  150 &
  \checkmark &
   &
   &
  YES &
  YES &
   &
  \checkmark &
   &
   &
   &
  Free &
   &
   &
   \\
Mapillary Vistas\cite{neuhold2017mapillary} &
  2017 &
  25k &
  - &
   &
  224 &
  \checkmark &
  \multicolumn{1}{c}{\checkmark} &
   &
  YES &
  YES &
   &
  \checkmark &
   &
   &
   &
  Partial Free &
   &
   &
   \\
Indian Driving Dataset (IDD)\cite{varma2019idd} &
  2018 &
  10k &
  - &
   &
  34 &
  \checkmark &
   &
   &
  YES &
  NO &
   &
  \checkmark &
   &
   &
   &
  Free &
   &
   &
   \\
Cityscapes\cite{cordts2016cityscapes} &
  2016 &
  25k &
  - &
   &
  30 &
  \checkmark &
   &
   &
  YES &
  NO &
   &
  \checkmark &
   &
   &
   &
  Free &
   &
   &
   \\
BDD100K\cite{yu2020bdd100k} &
  2018 &
  100M &
  100k &
   &
  55 &
  \checkmark &
  \multicolumn{1}{c}{\checkmark} &
  \multicolumn{1}{c}{\checkmark} &
  YES &
  YES &
   &
  \checkmark &
   &
   &
   &
  Free &
   &
   &
   \\
KITTI\cite{geiger2012we} &
  2012 &
  12k &
  - &
   &
  11 &
  \checkmark &
   &
   &
  YES &
  NO &
   &
  \checkmark &
  \multicolumn{1}{c}{\checkmark} &
   &
  \multicolumn{1}{c}{\checkmark} &
  Free &
   &
   &
   \\
ApolloScape\cite{huang2018apolloscape} &
  2018 &
  140k &
  73 &
   &
  28 &
  \checkmark &
  \multicolumn{1}{c}{\checkmark} &
  \multicolumn{1}{c}{\checkmark} &
  YES &
  YES &
   &
  \checkmark &
  \multicolumn{1}{c}{\checkmark} &
   &
  \multicolumn{1}{c}{\checkmark} &
  Free &
   &
   &
   \\
WoodScape\cite{yogamani2019woodscape} &
  2019 &
  10k &
  - &
   &
  40 &
  \checkmark &
   &
   &
  YES &
  YES &
   &
  \multicolumn{1}{l}{} &
   &
  \multicolumn{1}{c}{\checkmark} &
   &
  Free &
   &
   &
   \\
Waymo Open Dataset\cite{sun2020scalability} &
  2019 &
  200k &
  1.95k &
   &
  4 &
  \checkmark &
  \multicolumn{1}{c}{\checkmark} &
  \multicolumn{1}{c}{\checkmark} &
  YES &
  YES &
   &
  \checkmark &
   &
   &
  \multicolumn{1}{c}{\checkmark} &
  Free &
   &
   &
   \\
nuScenes\cite{caesar2020nuscenes} &
  2019 &
  1.4M &
  1k &
   &
  23 &
  \checkmark &
  \multicolumn{1}{c}{\checkmark} &
  \multicolumn{1}{c}{\checkmark} &
  YES &
  YES &
   &
  \checkmark &
   &
   &
  \multicolumn{1}{c}{\checkmark} &
  Free &
   &
   &
   \\
Lyft Level 5\cite{houston2021one} &
  2019 &
  55k &
  175k &
   &
  366 &
  \checkmark &
  \multicolumn{1}{c}{\checkmark} &
  \multicolumn{1}{c}{\checkmark} &
  YES &
  YES &
   &
  \multicolumn{1}{c}{\checkmark} &
   &
   &
  \multicolumn{1}{c}{\checkmark} &
  Free &
   &
   &
   \\ \cline{1-17}
\end{tabular}}
\caption{A summary of dataset in autonomous driving}
\label{table:dataset}
\end{table}
\vspace{-20pt}

\textbf{Tasks Annotation}. Object detection, whether in 2D or 3D format, fundamentally involves identifying each object within an image or scene and accurately representing it with a bounding box and an associated label. For 2D object detection, widely used datasets include COCO, Mapillary Vistas, IDD, Cityscapes, BDD100K, KITTI, Waymo Open Dataset, ApolloScape, and WoodScape. Among these, KITTI, Waymo Open Dataset, nuScenes, and ApolloScape extend their applicability to 3D object detection, offering a rich spatial data source for more complex analyses. Regarding semantic segmentation, the task shifts to identifying and delineating objects of specific classes at the pixel level within an image. This requires a different approach compared to instance segmentation, where the focus is on classifying each pixel. The 12 datasets mentioned are well-equipped with labels and annotations tailored for semantic and instance segmentation tasks, providing a foundation for training and testing segmentation algorithms. BDD100K stands out as a particularly versatile public dataset, encompassing annotations not only for standard object detection but also for lane segmentation and drivable area segmentation. However, it's important to note that lane and drivable area segmentation data can also be derived from other types of annotations, such as those found in semantic or instance segmentation datasets. Datasets like Mapillary Vistas, IDD, Cityscapes, ADE20K, KITTI, ApolloScape, Waymo Open Dataset, and nuScenes, for example, can be repurposed to extract valuable lane and drivable area information through appropriate methodologies. For depth estimation tasks, Woodscape emerges as a notable dataset. It provides a platform to train and test algorithms, facilitating the development of models capable of interpreting and reconstructing the depth of a scene from fish-eye camera data.

\textbf{Size}. The COCO dataset contains more than 200,000 labeled images and is mainly used for object detection, image segmentation, and key point detection tasks but does not cover depth estimation tasks. In the ADE20K dataset, there are more than 25,000 training images and 2,000 validation images, and it is mainly suitable for semantic and instance segmentation but does not involve object detection and depth estimation tasks. IDD includes 10,000 images and is mainly applied to semantic segmentation tasks, not involving object detection and depth estimation. Likewise, the Mapillary Vistas dataset, consisting of 25,000 images, is also primarily used for image segmentation tasks. The Cityscapes dataset has 5,000 high-quality images with detailed annotations and 20,000 images with coarse annotations, providing rough and incomplete polygonal annotations. The ApolloScape dataset contains more than 140,000 images and corresponding 3D LiDAR point clouds, which can support image segmentation, object detection, and depth estimation tasks. The KITTI dataset has more than 12,000 images; similarly, it can perform object detection, image segmentation, and depth estimation tasks. Waymo Open Dataset consists of 1,950 high-quality videos (20s duration each) captured by full sensors. Similarly, the nuScenes dataset includes 1,000 videos (20s duration each), supporting object detection, segmentation, and depth estimation tasks. Lyft Level 5 dataset includes 170,000 scenes, each 25 seconds long, capturing the movement of the self-driving vehicle, traffic participants around it, and the traffic lights state. It can be used for object detection and segmentation. The BDD100K dataset contains 100,000 videos (40 seconds each), supporting ten tasks, including object detection, drivable area segmentation, lane segmentation, etc. The WoodScape dataset contains over 10,000 wide-angle camera images and supports nine tasks, including object detection, instance segmentation, depth estimation tasks, etc.

\textbf{Diversity}. The COCO dataset contains 80 kinds of object categories and covers a variety of scenes, including indoor and outdoor environments, but is relatively limited in the diversity of geographical locations and environmental conditions, mainly focusing on general weather and daytime conditions. The ADE20K dataset covers 150 categories, with about 25k images, including various urban and rural scenes, but most of these images are collected under sunny and daytime conditions. Datasets focused on autonomous driving, such as ApolloScape, Waymo Open Dataset, nuScenes, BDD100K, Lyft Level 5, and WoodScape, contain richer environments and geographic conditions. For example, the nuScenes dataset contains 23 categories and 1000 scenes, covering the cities of Boston and Singapore. They contain various road types and driving conditions, such as urban, suburban, night and day, sunny, rainy snow, etc. The BDD100K dataset is even larger, containing as many as 100,000 videos, covering multiple USA cities, including multiple weather and time conditions, such as sunny, cloudy, rainy, dusk, night, etc. Urban street view-focused datasets, such as Mapillary Vistas and Cityscapes, cover multiple cities and street environments. The Cityscapes dataset contains 30 categories, covering street views of 50 cities. The Mapillary Vistas dataset covers 6 continents worldwide, including 124 semantic object categories and 100 instance-specifically annotated categories, reflecting a more comprehensive geographical coverage and environmental diversity. Similarly, the IDD dataset focuses on the road environment in India, which contains 15 categories, covering 2 cities in India and various unique traffic environments.
\begin{table}[!htb]
\centering
\begin{tabular}{@{}lccccccc@{}}
\toprule
Dataset                      & \multicolumn{7}{c}{Tasks}                                                                \\ \midrule
                             & OD         & OD(3D)     & IS         & SS         & LS         & DAS        & DE         \\ \midrule
COCO                         & \checkmark &            & \checkmark & \checkmark &            &            &            \\
ADE20K                       &            &            & \checkmark & \checkmark & $\ast$     & $\ast$     &            \\
Mapillary Vistas             & \checkmark &            & \checkmark & \checkmark & $\ast$     & $\ast$     &            \\
Indian Driving Dataset (IDD) & \checkmark &            & \checkmark & \checkmark &            &            &            \\
Cityscapes                   & \checkmark &            & \checkmark & \checkmark & $\ast$     & $\ast$     &            \\
BDD100K                      & \checkmark &            & \checkmark & \checkmark & \checkmark & \checkmark &            \\
KITTI                        & \checkmark & \checkmark & \checkmark & \checkmark & $\ast$     & $\ast$     &            \\
ApolloScape                  & \checkmark & \checkmark & \checkmark & \checkmark & $\ast$     & $\ast$     &            \\
WoodScape                    & \checkmark & \checkmark & \checkmark & \checkmark &            &            & \checkmark \\
Waymo Open Datase            &            & \checkmark & \checkmark & \checkmark & $\ast$     & $\ast$     &            \\
nuScene                      &            & \checkmark & \checkmark & \checkmark & $\ast$     & $\ast$     &            \\
Lyft Level 5                 &            & \checkmark & \checkmark & \checkmark &            &            &            \\ \bottomrule
\end{tabular}
\caption{Tasks supported by different datasets. $"\ast"$ indicates that the output of this task can be obtained indirectly}
\label{dataset tasks}
\end{table}
\vspace{-20pt}

\textbf{Limitation}. COCO is a large dataset focused on common objects, and its richness and diversity make it a benchmark in the field of object detection and segmentation. However, since COCO focuses mainly on everyday objects, there may be some limitations for specific application domains, such as autonomous driving. Similarly, ADE20K covers many scenes and object categories, but its generality may make it imprecise on specific tasks. For tasks related to autonomous driving, ApolloScape and Waymo Open Dataset provide many 2D and 3D labels for autonomous driving applications, including various road conditions and traffic elements, which are very valuable for the research and development of autonomous driving. However, these datasets are mainly focused on specific cities or regions and may have limitations for tasks that require broad geographic coverage or specific environments. In the study of urban environment and road understanding, Mapillary Vistas and Cityscapes provide many street-level images. However, since they are mainly concentrated in urban settings, there may be some limitations in terms of environmental diversity. For tasks that require understanding a specific geographic or cultural environment, such as autonomous driving in an Indian environment, IDD provides a large amount of data with geographic characteristics, which is very valuable for related research. However, since IDD is mainly focused on specific environments in India, there may be some limitations for tasks that require extensive geographic coverage or diverse environments. BDD100K, Lyft Level 5, and WoodScape each provide various autonomous driving-related data from different perspectives. BDD100K provides rich driving images in time and weather conditions, but its annotations may not be as fine-grained as other datasets. Lyft Level 5 provides a large amount of 3D data related to autonomous driving, but its geographic coverage may be limited. WoodScape provides on-board camera data from multiple perspectives, increasing the richness of the data, but the data scale may be relatively limited. Overall, these datasets have their own advantages and disadvantages and are suitable for different tasks and research needs. The choice of which dataset to use depends on research goals, mission requirements, and specific needs for data diversity, annotation quality, and scale.

\vspace{-5pt}
\subsection{Evaluation Metrics}
\vspace{5pt}
\subsubsection{Performance Metrics}
\begin{figure}[!htb]
  \centering
  \includegraphics[scale=0.4]{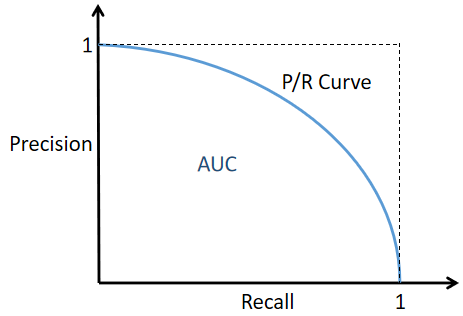}
  \caption{P/R curve}
  \label{fig:PRcurve}
\end{figure}

\ 
\newline
\textbf{Object detection}. For object detection tasks, indicators such as Precision, Recall, Accuracy, error rate, F-1 score, and Average Precision (AP) are usually used for evaluation. Precision is the proportion of correct objects among detected objects and recall is the proportion of correctly detected objects to all real objects. For object detection, we usually define a threshold (e.g., IoU > 0.5), and if the Intersection over Union (IoU) of the predicted bounding box and the ground-truth bounding box is greater than this threshold, we consider the prediction to be correct. Based on this, precision, recall, accuracy, and error rate can be defined as:

\begin{equation}
Precision =\frac{TP}{TP + FP}
\label{Precision}
\end{equation}

\begin{equation}
Recall =\frac{TP}{TP + FN}
\label{Recall}
\end{equation}

\begin{equation}
Accuracy =\frac{TP+TN}{TP + FP + TN + FN}
\label{accuracy}
\end{equation}

\begin{equation}
Error Rate = 1-Accuracy
\label{error}
\end{equation}
where TP (True Positive) is the number of correctly predicted objects, FP (False Positive) is the number of incorrectly predicted objects, and FN (False Negative) is the number of true objects not detected.
Usually, we use precision and recall to measure the quality of the model but to weigh these two quantities at the same time, it needs to be calculated by the following F-score,
\begin{equation}
F-score =\frac{(1+\beta^2)\times Precision\times Recall }{\beta^2\times Precision + Recall}
\label{F-score}
\end{equation}
The F1-score is calculated when $\beta=1$. This score gives equal importance to both precision and recall. However, in some cases, accuracy may be more important than recall. In such cases, we can adjust the value of $\beta$ to be less than 1. Similarly, if we think that the recall rate is more important, we can adjust the value of $\beta$ to be greater than 1.
To evaluate the performance of a model comprehensively, we use the area under the recall and precision curve, known as the AP.
\begin{equation}
AP=\int P(r)dr
\label{AP}
\end{equation}
where P(r) is the P/R curve. Most of the time, algorithms must have high precision and high recall. However, most machine learning algorithms usually involve a trade-off between the two. The curve in Figure~\ref{fig:PRcurve} represents the P/R curve, and the area under the curve (AUC) represents the AP. Generally, a good PR curve has a larger AUC.

For object detection and instance segmentation tasks, the COCO dataset introduces more complex evaluation indicators, such as $AP@[.5:.05:.95]$, that is, at different IoU thresholds (from 0.5 to 0.95, step is 0.05 ) to calculate the value of AP. This metric encourages models to maintain high performance under stricter IoU thresholds.
\vspace{-10pt}
\vspace{-12pt}
The object detection model usually uses the mAP to describe the performance. The higher the mAP value, the better the detection result of the object detection model on a given dataset. Corresponding to different IoU thresholds, mAP50 is a commonly used metric.
\begin{equation}
mAP=\frac{\sum_{i=1}^{C}AP_{i}}{C} 
\label{mAP}
\end{equation}
where C is the number of classes.
Waymo Open Dataset introduces task-specific metrics such as average precision weighted by head (APH), which evaluates the accuracy of detecting the heading of the object,
\begin{equation}
APH=\int H(r)dr
\label{APH}
\end{equation}
where H(r) is computed similar to P(r), but each true positive is weighted by heading accuracy defined as $min(| \widetilde{\theta} - \theta |, 2\pi - | \widetilde{\theta} - \theta |)/\pi$, where $\widetilde{\theta}$ and $\theta$ are the predicted heading and the ground truth heading in radians within $[-\pi, \pi]$.

In addition to the above common performance metrics, some datasets also refer to other evaluation indicators. The nuScenes dataset introduces composite metrics such as NuScenes Detection Score (NDS), where NDS considers different types of objects and detected attributes (such as speed, location, etc.). It can be calculated by following the equation,
\begin{equation}
 NDS=\frac{1}{10}[5mAP + \sum_{mTP}(1-min(1,mTP))]
 \label{NDS}
\end{equation}
For each TP metric, we compute the mTP over all classes by following the equation,
\begin{equation}
mTP=\frac{\sum_{i=1}^{C}TP_{i}}{C} 
\label{mTP}
\end{equation} 
where TP refers to various indicators, which are explained in detail in the nuScenes dataset\cite{caesar2020nuscenes}. 

\textbf{Segmentation}. Segmentation is the classification at the pixel level. Therefore, it is usually evaluated using indicators such as Pixel Accuracy and Mean Intersection over Union (mIoU). Pixel accuracy is defined as the proportion of correct predictions among all pixels, specifically,
\begin{equation}
Pixel Accuracy =\frac{p_{pred}}{p_{all}}
\label{pixel accuracy}
\end{equation}
where, $p_{pred}$ is the number of predicted correct pixels and $p_{all}$ is the number of all pixels.
mIoU calculated by the following equation,
\begin{equation}
mIoU = \frac{I}{U \times C}
\label{mIoU}
\end{equation}
where I is the intersection of predicted area and real area, U is the union of predicted area and real area, and C is the number of classes.
Panoptic Segmentation\cite{kirillov2019panoptic} considers both semantic segmentation and instance segmentation tasks, while Panoptic Quality (PQ) is used to evaluate the performance of the model on the panoptic segmentation task.
The calculation of PQ includes two parts: segmentation quality (SQ) and recognition quality (RQ). PQ can be written as the following formula,
\begin{equation}
\begin{aligned}
PQ & = \underbrace{\frac{\sum_{(p,g)\in TP} IoU(p,g)}{|TP|}}_{segmentation quality (SQ)} \times \underbrace{\frac{|TP|}{|TP|+\frac{1}{2}|FP|+\frac{1}{2}|FN|}}_{recognition quality (RQ)}
& = \frac{\sum_{(p,g)\in TP} IoU(p,g)}{|TP|+\frac{1}{2}|FP|+\frac{1}{2}|FN|}
\end{aligned}
\label{PQ}
\end{equation}
where RQ is the familiar F1 score widely used for quality estimation in detection settings. SQ is simply the average IoU of matched segments. In equation~\ref{PQ}, $\frac{\sum_{(p,g)\in TP} IoU(p,g)}{|TP|}$ is simply the average IoU of matched segments, while $\frac{1}{2}|FP|+\frac{1}{2}|FN|$ is added to the denominator to penalize segments without matches.


\textbf{Depth estimation}. The evaluation indicators of depth estimation include mean relative error (MRE) and mean logarithmic error (MLE). Among them, the MRE is the average value of the relative error between the predicted depth value and the real depth value calculated for all pixels, that is, 
\begin{equation}
 MRE=\frac{1}{N}\sum \frac{|d_{pred} - d_{true}|}{d_{true}}   
\label{Mean Relative Error}
\end{equation}
where$d_{pred}$ is the predicted depth, $d_{true}$ is the ground truth depth and N is the total number of pixels. The MLE is the average of the logarithmic difference between the predicted depth value and the true depth value for all pixels, 
\begin{equation}
 MLE=\sum \frac{|\log d_{pred} - \log d_{true}|}{N}
 \label{MLE}
\end{equation}
\vspace{-25pt}
\subsubsection{Efficiency Metrics}
\ 
\newline
In machine learning model efficiency evaluation, various metrics are employed to assess the computational intensity and the storage requirements. The number of parameters, indicating the count of elements within the model's weight tensors, is a key metric reflecting model complexity. A higher parameter count often suggests a more sophisticated model with enhanced learning capabilities but also increases computational and memory demands. The overall storage footprint of a model, an important consideration for deployment, can be calculated by multiplying these parameters by the storage size per parameter, typically in bytes, giving a clear picture of the model's size.

Furthermore, the efficiency of a model during operation is largely gauged by its computational complexity. Multiply-accumulate operations (MAC) play a central role as the most common calculation in neural networks, serving as a primary indicator of computational load. Similarly, the Number of Operations (OP) or Floating Point Operations (FLOP) provide insights into the total computational work required by the model. These figures, especially when converted into OPS (Operations Per Second) or FLOPS (Floating Point Operations Per Second), offer a dynamic view of the model's processing speed and efficiency.

Responsiveness is another important efficiency indicator of neural network models. Real-time tasks can only be achieved with high computation speed. Frames per second (FPS) can be used to evaluate the computation speed of a perception task. In image processing tasks, it represents the number of images that can be processed per second. It is worth noting that valid FPS comparisons should be made on identically configured hardware. Latency can also be used to evaluate model computing performance, which is the reciprocal of FPS. Another metric related to latency is throughput. Under the same bandwidth, if the delay is higher, the image processing is often slower, and the throughput is lower.

In addition to these computational metrics, basic energy units like watts or kilowatts are instrumental in assessing the model's power consumption or utilization. This aspect is particularly critical in scenarios where energy efficiency is paramount, such as in mobile or edge systems. By collectively analyzing these metrics, one can comprehensively understand a model's efficiency, balancing its performance and practical deployability.

\section{Techniques for Panoptic Perception} \label{Techniques for Panoptic Perception}

Accurate perception of the surrounding environment is crucial for autonomous driving, and this is where panoptic perception comes into play. It involves tasks such as object detection, semantic segmentation, lane segmentation, drivable area segmentation, depth estimation, and instance segmentation. To achieve this efficiently and accurately, researchers have developed various networks and technologies based on images, LiDAR point clouds, or a fusion of both. In this section, we categorize these technologies based on the input type and provide an overview of their architectures, advantages, and limitations. This section aims to help readers gain a deeper understanding of the current state of panoptic perception technology and serve as a reference for future research and development.
\vspace{-15pt}
\begin{figure}[!htb]
  \centering
  \includegraphics[scale=0.45]{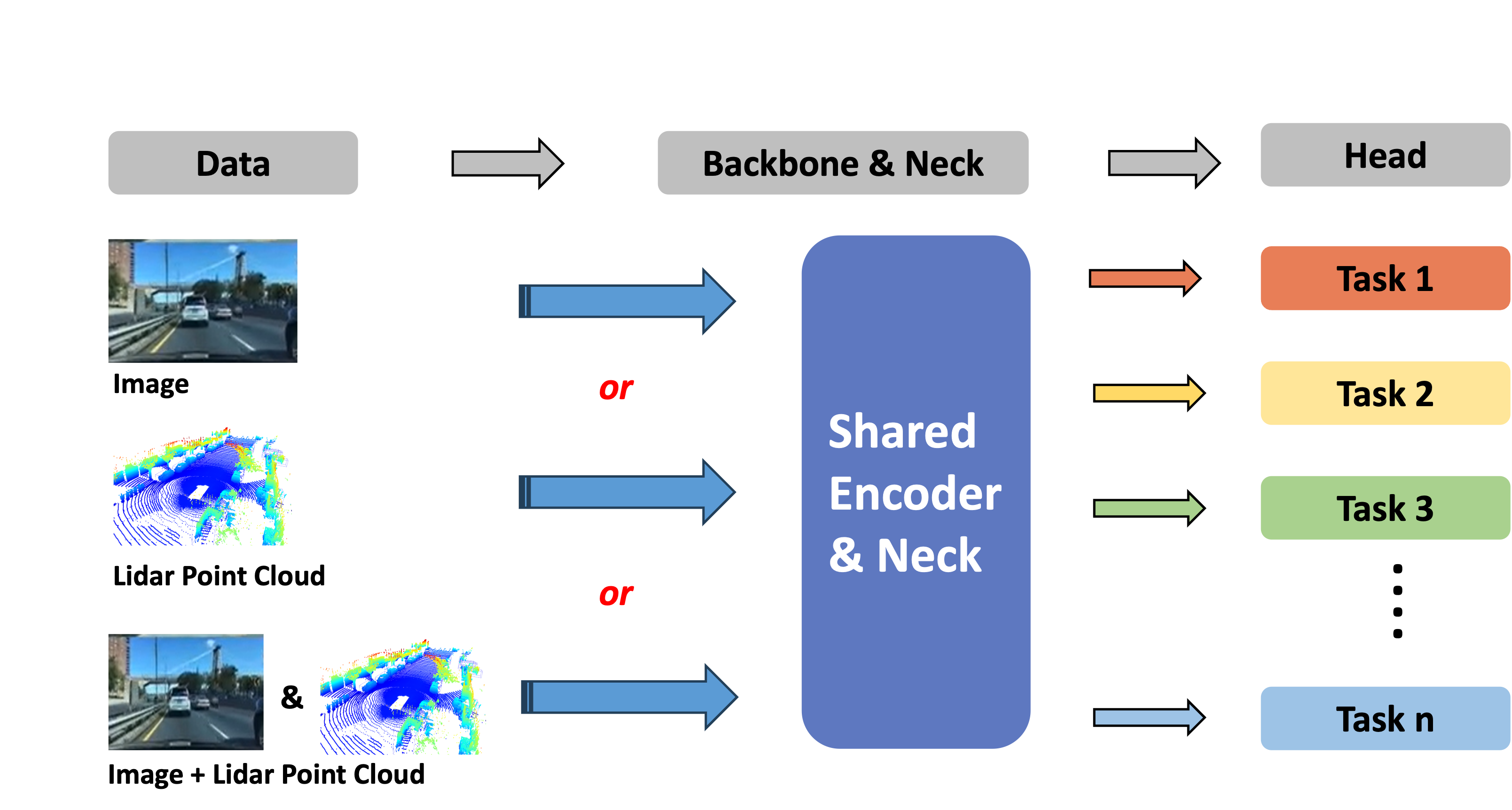}
  \caption{Panoptic perception network architecture }
  \label{fig:Multi-task Network Architecture}
\end{figure}
\vspace{-15pt}
\subsection{Overview of Current Techniques for Panoptic Perception}
In autonomous driving panoptic perception, a prevalent structure is a multi-task network, depicted in figure~\ref{fig:Multi-task Network Architecture}. This architecture comprises three integral parts: the backbone, neck, and head. The backbone, typically a deep network like ResNet~\cite{he2015deep} or VGG~\cite{simonyan2015deep}, is tasked with initial feature extraction from various input sources, such as images, LiDAR point clouds, or their fusion. This stage is critical for effectively processing the high-dimensional and complex data characteristic of autonomous driving scenarios. Following the backbone is the neck, a component dedicated to further processing and refining these extracted features. This may involve advanced techniques such as cross-layer feature fusion, enhancement, and selection tailored to enrich the feature quality. The neck's design is pivotal, requiring careful consideration to accommodate the diverse range of tasks and to provide task-appropriate feature inputs. The final segment of the architecture is the head, comprising multiple sub-networks, each specialized for a distinct task. These sub-networks are meticulously tailored, incorporating task-specific network structures and loss functions to optimize performance. This multi-task network architecture leverages a shared backbone and neck for efficient feature extraction, enhancing learning efficiency and generalization across tasks. Concurrently, it ensures effective task-specific processing through its specialized heads.

\vspace{-10pt}
\subsection{Image-based Network}
Panoptic perception techniques are utilized in autonomous driving to perform various tasks discussed in section~\ref{Panoptic Perception Tasks} by using images as input. Images are advantageous for processing high-level visual information due to their rich visual content, such as color, texture, and shape of the environment.

\vspace{-10pt}
\subsubsection{Backbone}
\ 
\newline
In the context of multi-task networks for autonomous driving perception, a shared encoder typically serves as the backbone, tasked with extracting both low-level and high-level features from images. A diversity of deep neural network architectures have been developed for this purpose, each with distinct characteristics.

\textbf{VGG}~\cite{simonyan2015deep}, a deep CNN architecture proposed by the Visual Geometry Group at Oxford University in 2014, exemplifies simplicity and efficacy in feature extraction. Characterized by its uniform use of 3x3 convolution kernels and ReLU activation functions, VGG has been utilized in various configurations, such as MultiNet's~\cite{teichmann2018multinet} use of the VGG network, DLT-Net's~\cite{qian2019dlt} implementation of VGG16, and CP-MLT's~\cite{chen2018multi} adoption of Oxford VGG. While its depth and straightforward structure contribute to effective feature extraction, the extensive depth and fully connected layers of VGGNet result in a relatively large computational footprint and parameter size, posing challenges in resource-limited autonomous driving systems.

\textbf{ResNet}~\cite{he2015deep}, introduced by Microsoft Research in 2015, is a deep residual network that addresses the difficulty of training deep neural networks. Incorporating a residual structure, ResNet facilitates the construction of extremely deep networks (e.g., ResNet50, ResNet101, ResNet152), and has demonstrated robust performance across multiple tasks in autonomous driving applications, as evidenced in architectures like MGNet~\cite{schon2021mgnet}, CERBERUS~\cite{scribano2022cerberus}, Sparse U-PDP~\cite{wang2023sparse}, MatrixVT~\cite{zhou2022matrixvt}, EfficientPS~\cite{mohan2021efficientps}, PETRv2~\cite{liu2022petrv2}, BEVFormer~\cite{li2022bevformer}, and BEVFormer v2~\cite{yang2023bevformer}. Despite its benefits in capturing detailed environmental and object features, ResNet shares a similar challenge with VGG regarding substantial computational resource requirements.

\textbf{EfficientNet}~\cite{tan2020efficientnet}, a network architecture proposed by Google in 2019, represents a significant advancement in balancing performance with computational efficiency. By simultaneously scaling network depth, width, and resolution, EfficientNet achieves enhanced performance with reduced computational demands, making it a suitable candidate for resource-constrained environments, as seen in its adoption in HybridNet~\cite{vu2022hybridnets}, EfficientPS~\cite{mohan2021efficientps}, and MGNet~\cite{schon2021mgnet}.

\textbf{CSP-Darknet}~\cite{bochkovskiy2020yolov4} offers a lightweight and efficient solution based on Darknet, with Cross Stage Partial (CSP) connections designed to enhance network efficiency. These connections facilitate a more direct gradient flow between feature maps, reducing computational complexity while preserving accuracy. This architecture is particularly advantageous for real-time autonomous driving systems, as demonstrated by its use in YOLOP~\cite{wu2022yolop}, CenterPNets~\cite{chen2023centerpnets}, and Sparse U-PDP~\cite{wang2023sparse}.
\begin{table}[!htb]
\begin{tabular}{lllcc}
\hline
Input Modality & Model basis       & Techniques     & Backbone                                                             & Neck                 \\ \hline
Image          & CNN-based         & YOLOP~\cite{wu2022yolop}                   & CSPDarknet~\cite{bochkovskiy2020yolov4}                                                          & FPN~\cite{lin2017feature}, SPP~\cite{he2015spatial}             \\ \cline{3-5} 
               &                   & YOLOPv2~\cite{han2022yolopv2}                 & E-ELAN~\cite{wang2022yolov7}                                                               & FPN, SPP             \\ \cline{3-5} 
               &                   & HybridNets~\cite{vu2022hybridnets}              & EfficientNet-B3~\cite{tan2020efficientnet}                                                      & BiFPN~\cite{tan2020efficientdet}                \\ \cline{3-5} 
               &                   & DLT-Net~\cite{qian2019dlt}                 & VGG16~\cite{simonyan2015deep}                                                                & FPN                  \\ \cline{3-5} 
               &                   & MultiNet~\cite{teichmann2018multinet}                & \begin{tabular}[c]{@{}c@{}}VGG~\cite{simonyan2015deep}\\ ResNet-50~\cite{he2015deep}\\ ResNet-101~\cite{he2015deep}\end{tabular} & N/A                  \\ \cline{3-5} 
               &                   & EfficientPS~\cite{mohan2021efficientps}             & EfficientNet~\cite{tan2020efficientnet}                                                        & FPN                  \\ \cline{3-5} 
               &                   & CenterPNets~\cite{chen2023centerpnets}             & CSPDarkNet~\cite{bochkovskiy2020yolov4}                                                           & FPN                  \\ \cline{3-5} 
 &  & MGNet~\cite{schon2021mgnet}        & \begin{tabular}[c]{@{}c@{}}ResNet-18~\cite{he2015deep}\\ EfficientNetLite0~\cite{tan2020efficientnet}\\ MNASNet100~\cite{tan2019mnasnet}\\ MobileNetV3~\cite{howard2019searching}\end{tabular} & GCM        \\ \cline{3-5} 
               &                   & ShuDA-RFBNet~\cite{wang2019shuda}            & ShuffleNet V2~\cite{ma2018shufflenet}                                                        & N/A                  \\ \cline{3-5} 
               &                   & AdvNet~\cite{liu2019advnet}                  & Enet~\cite{paszke2016enet}                                                                 & N/A                  \\ \cline{3-5} 
               &                   & CERBERUS~\cite{scribano2022cerberus}                & \begin{tabular}[c]{@{}c@{}}ResNet~\cite{he2015deep} \\ Mobilenet~\cite{howard2017mobilenets}\end{tabular}          & BiFPN                \\ \cline{3-5} 
               &                   & Sparse U-PDP~\cite{wang2023sparse}            & \begin{tabular}[c]{@{}c@{}}ResNet-50~\cite{he2015deep}\\  CSPDarknet~\cite{bochkovskiy2020yolov4}\end{tabular}      & N/A                  \\ \cline{3-5} 
               &                   & $M^{2}BEV$~\cite{xie2022m} & ResNeXt-101~\cite{xie2017aggregated}                                                          & FPN                  \\ \cline{3-5} 
               &                   & Omnidet~\cite{kumar2021omnidet}                 & SAN~\cite{zhao2020exploring}                                                                  & PAC~\cite{su2019pixel}                  \\ \cline{3-5} 
               &                   & CP-MTL~\cite{chen2018multi}                  & VGG~\cite{simonyan2015deep}                                                           & N/A                  \\ \cline{3-5} 
 &  & MatrixVT~\cite{zhou2022matrixvt}     & \begin{tabular}[c]{@{}c@{}}ResNet-50~\cite{he2015deep}\\ ResNet-101~\cite{he2015deep} \\ VoVNet2-99~\cite{lee2020centermask}\end{tabular}                           & SECOND FPN~\cite{yan2018second} \\
               &                   &                         &                                                                      & \multicolumn{1}{l}{} \\ \cline{2-5} 
               & Transformer-based & BEVerse~\cite{zhang2022beverse}                 & Swin-transformer~\cite{liu2021swin}                                                     & N/A                  \\
               &                   &                         &                                                                      & \multicolumn{1}{l}{} \\ \cline{2-5} 
               & CNN\&Transformer  & VE-Prompt~\cite{liang2023visual}               & Swin-transformer~\cite{liu2021swin}                                                     & FPN                  \\ \cline{3-5} 
 &  & PETRv2~\cite{liu2022petrv2}       & \begin{tabular}[c]{@{}c@{}}ResNet~\cite{he2015deep}\\ VoVNetV2~\cite{lee2020centermask}\\ EfficientNet~\cite{tan2020efficientnet}\end{tabular}                         & N/A        \\ \cline{3-5} 
               &                   & BEVFormer~\cite{li2022bevformer}               & \begin{tabular}[c]{@{}c@{}}ResNet-101~\cite{he2015deep}\\ VoVnet-99~\cite{lee2019energy}\end{tabular}         & FPN                  \\ \cline{3-5} 
 &  & BEVFormer v2~\cite{yang2023bevformer} & \begin{tabular}[c]{@{}c@{}}ResNet~\cite{he2015deep}\\ DLA~\cite{yu2019deep}\\ VoVNet~\cite{lee2019energy}\\ InternImage~\cite{wang2023internimage}\end{tabular}                      & N/A        \\ \hline
\end{tabular}
\caption{Backbone and neck for image-based models}
\label{table:Backbone and Neck for Image-based Models}
\end{table}
\vspace{-20pt}
Other noteworthy architectures include \textbf{ResNeXt-101}, used in $M^2BEV$, which incorporates group convolution based on ResNet to improve accuracy without substantially increasing model complexity. \textbf{ShuffleNet V2}~\cite{ma2018shufflenet}, utilized in ShuDA-RFBNet~\cite{wang2019shuda}, aims to minimize computational requirements while maintaining high accuracy. 

\textbf{MobileNet}~\cite{howard2017mobilenets}, with its depthwise separable convolutions, significantly reduces both computational load and parameter size, proving effective in mobile and edge computing applications, as validated in MGNet~\cite{schon2021mgnet} and CERBERUS~\cite{scribano2022cerberus}.

\textbf{VoVNet}~\cite{lee2019energy}, implemented in models like MatrixVT, PETRv2, BEVFormer, and BEVFormerv2, introduces the OSA module (One-Shot Aggregated module) for efficient feature fusion and reduced computational demand.

Distinct from the aforementioned CNN-based architectures, the \textbf{Swin Transformer}~\cite{liu2021swin}, as utilized in BEVerse, exemplifies the emerging trend of transformer-based backbones in visual computing tasks. This model stands out for its adeptness in global context comprehension and detailed feature extraction, particularly relevant in the complex scenarios of autonomous driving perception. At the heart of the Swin Transformer's strengths is its global self-attention mechanism, a key attribute of transformer models. This mechanism allows for an extensive and nuanced analysis of the entire image, enabling the network to capture and integrate contextual information from a wide field of view. Unlike traditional CNNs that primarily operate within local receptive fields, the Swin Transformer processes visual data in a more holistic manner. This global processing capability not only enhances the accuracy of feature extraction but also significantly improves model interpretability. Such attributes make the Swin Transformer particularly suitable for tasks where understanding the broader context and intricate details of a scene is crucial. Transformer-based models of visual computing, such as the BEVerse~\cite{zhang2022beverse}, have great potential for autonomous driving by providing comprehensive image understanding.

The selection of a backbone architecture for autonomous driving perception tasks should be informed by careful consideration of the specific application requirements, hardware resource constraints, and empirical performance evaluations. Each architecture presents a unique set of advantages and trade-offs, necessitating a tailored approach to optimize performance in the diverse landscape of autonomous driving technologies.
\subsubsection{Neck}
\ 
\newline
In multi-task network architectures for autonomous driving perception, the neck plays a crucial role in enhancing features extracted by the backbone. The Feature Pyramid Network (FPN)~\cite{lin2017feature} is a commonly employed neck structure, known for its multi-scale feature extraction capabilities. FPN achieves this by integrating a top-down architecture with horizontal connections, allowing the simultaneous processing of both high-level and low-level features, each carrying distinct scales and semantic information. This design makes FPN particularly adept at recognizing and localizing objects of varying sizes in autonomous driving tasks. However, one limitation of FPN is the relative simplicity of its feature fusion approach, which might lead to potential information loss. To address this, the Bi-directional Feature Pyramid Network (BiFPN)~\cite{tan2020efficientdet} introduces a more sophisticated feature fusion system with both top-down and bottom-up information flows. In autonomous driving perception tasks, BiFPN enhances performance by providing richer and more accurate feature representation. This increased accuracy, however, comes at the cost of higher computational complexity, impacting both computational cost and processing time. Additionally, for processing vast quantities of sparse 3D point cloud data, architectures like SECOND FPN~\cite{yan2018second} employ Sparse CNNs to construct feature pyramids effectively. For instance, MatrixVT utilizes SECOND FPN to augment feature extraction in the image and BEV analysis.

In specific visual tasks like semantic segmentation, object detection, and recognition, the integration of global context information is crucial. It enables the network to comprehend better and interpret the entire scene and the interrelations between various elements within the image. For example, YOLOP and YOLOPv2, alongside using FPN, incorporate the Spatial Pyramid Pooling (SPP)~\cite{he2015spatial} module to process input images of varying sizes, facilitating the extraction of more comprehensive context information. MGNet employs the Global Context Module (GCM)~\cite{schon2021mgnet}, which derives a global feature descriptor by performing pooling operations across the entire feature map, encapsulating contextual details of the entire image. Omnidet adopts pixel-adaptive convolution (PAC)~\cite{kumar2021omnidet} to infuse semantic knowledge extracted from features into distance estimation, breaking the spatial invariance of conventional convolutions. This approach allows for the integration of location-specific semantic knowledge into multi-level distance features, enhancing the model's accuracy in complex perception tasks.

The necessity and choice of a neck in a multi-task network are contingent upon the specific requirements of the feature extraction process. While FPN and its variants are beneficial for tasks demanding multi-scale feature extraction and high spatial resolution, they may be superfluous or even counterproductive for simpler tasks or when working with data that has already been preprocessed or has high-level features. The decision to implement a neck and the type of neck to be used must, therefore be tailored to the specific demands of the task at hand and the characteristics of the data being analyzed.
\subsubsection{Head} \label{image-based head}
\ 
\newline
After the image features are processed through the backbone and further refined by the neck, the next step in the network for autonomous driving perception involves directing these enhanced image-based features toward various heads or task-specific decoders. This stage is essential for interpreting the image features for specialized tasks.
\begin{table}[!htb]
\begin{tabular}{lllccclll}
\hline
Input Modality &
  Model basis & Techniques
   & 
  \multicolumn{1}{l}{Tasks} &
  \multicolumn{1}{l}{} &
  \multicolumn{1}{l}{} &
   &
   &
   \\ \cline{4-9} 
 &
   &
   &
  \multicolumn{1}{l}{OD} &
  \multicolumn{1}{l}{LS} &
  \multicolumn{1}{l}{DAS} &
  IS &
  SS &
  DE \\ \hline
Image &
  CNN-based &
  YOLOP &
  \checkmark &
  \checkmark &
  \checkmark &
   &
   &
   \\
 &
   &
  YOLOPv2 &
  \checkmark &
  \checkmark &
  \checkmark &
   &
   &
   \\
 &
   &
  HybridNets &
  \checkmark &
  \checkmark &
  \checkmark &
   &
   &
   \\
 &
   &
  DLT-Net &
  \checkmark &
  \checkmark &
  \checkmark &
   &
   &
   \\
 &
   &
  MultiNet &
  \checkmark &
   &
   &
   &
  \multicolumn{1}{c}{\checkmark} &
   \\
 &  & EfficientPS & \multicolumn{1}{l}{} & \multicolumn{1}{l}{} & \multicolumn{1}{l}{} & \multicolumn{1}{c}{\checkmark} & \multicolumn{1}{c}{\checkmark} &                       \\
 &
   &
  CenterPNets &
  \checkmark &
  \checkmark &
  \checkmark &
   &
   &
   \\
 &  & MGNet       & \multicolumn{1}{l}{} & \multicolumn{1}{l}{} & \multicolumn{1}{l}{} & \multicolumn{1}{c}{\checkmark} & \multicolumn{1}{c}{\checkmark} & \multicolumn{1}{c}{\checkmark} \\
 &
   &
  ShuDA-RFBNet &
  \checkmark &
  \multicolumn{1}{l}{} &
  \checkmark &
   &
   &
   \\
 &
   &
  AdvNet &
  \checkmark &
  \checkmark &
  \multicolumn{1}{l}{} &
   &
   &
   \\
 &
   &
  CERBERUS &
  \checkmark &
  \checkmark &
  \multicolumn{1}{l}{} &
   &
   &
   \\
 &
   &
  Sparse U-PDP &
  \checkmark &
  \checkmark &
  \checkmark &
   &
   &
   \\
 &
   &
  $M^{2}BEV$ &
  \checkmark &
  \checkmark &
  \checkmark &
   &
  \multicolumn{1}{c}{\checkmark} &
   \\
 &
   &
  Omnidet &
  \checkmark &
  \multicolumn{1}{l}{} &
  \multicolumn{1}{l}{} &
   &
  \multicolumn{1}{c}{\checkmark} &
  \multicolumn{1}{c}{\checkmark} \\
 &
   &
  CP-MTL &
  \checkmark &
  \multicolumn{1}{l}{} &
  \multicolumn{1}{l}{} &
   &
   &
  \multicolumn{1}{c}{\checkmark} \\
 &
   &
  MatrixVT &
  \checkmark &
  \checkmark &
  \checkmark &
   &
  \multicolumn{1}{c}{\checkmark} &
   \\
 &
   &
   &
  \multicolumn{1}{l}{} &
  \multicolumn{1}{l}{} &
  \multicolumn{1}{l}{} &
   &
   &
   \\ \cline{2-9} 
 &
  Transformer-based &
  BEVerse &
  \checkmark &
  \checkmark &
  \multicolumn{1}{l}{} &
   &
  \multicolumn{1}{c}{\checkmark} &
   \\
 &
   &
   &
  \multicolumn{1}{l}{} &
  \multicolumn{1}{l}{} &
  \multicolumn{1}{l}{} &
   &
   &
   \\ \cline{2-9} 
 &
  CNN\&Transformer &
  VE-Prompt &
  \checkmark &
  \checkmark &
  \checkmark &
   &
  \multicolumn{1}{c}{\checkmark} &
   \\
 &
   &
  PETRv2 &
  \checkmark &
  \checkmark &
   &
   &
  \multicolumn{1}{c}{\checkmark} &
   \\
 &
   &
  BEVFormer &
  \checkmark &
  \checkmark &
   &
   &
  \multicolumn{1}{c}{\checkmark} &
   \\
 &
   &
  BEVFormer v2 &
  \checkmark &
  \checkmark &
   &
   &
  \multicolumn{1}{c}{\checkmark} &
   \\ \hline
\end{tabular}
\caption{Tasks head for image-based models}
\label{table:Tasks Head for Image-based Models}
\end{table}
\vspace{-18pt}

\textbf{Object Detection}. 
Once image features are extracted by the shared backbone and refined by the neck in the network, their subsequent processing is determined by the specific approach of the object detection head. In the models we surveyed, YOLOP and YOLOPv2 employ the YOLO method, utilizing One-stage Anchor-based methods for object detection. MultiNet\cite{teichmann2018multinet}, on the other hand, adopts a Two-stage Anchor-based approach, employing Regions of Interest (RoI) for classifying areas and refining bounding boxes. CenterPNets opts for a One-stage, Anchor-free method, directly regressing on key point heat maps, sizes, and offsets. This approach negates the need for predefined anchor box ratios determined by K-means clustering and the complexity of Non-Maximum Suppression (NMS) processing. Both BEVFormer and BEVFormer v2 process images from monocular cameras across multiple perspectives to obtain BEV features, enabling 3D object detection through an enhanced DETR method.

\textbf{Instance Segmentation}.
Networks processing image data, characterized by their ordered pixel arrangement and density, excel in instance segmentation tasks. Multi-task networks, in particular, leverage shared feature extraction layers, enabling different tasks to benefit from common low-level and mid-level features. For instance, features essential for object detection can also significantly enhance instance segmentation. This feature-sharing strategy alleviates the computational load and augments the network's capability to discern varied visual cues. Training multiple tasks simultaneously in an end-to-end framework allows the network to develop more generalized feature representations, bridging different tasks. Furthermore, understanding the interplay between tasks enables the network to interpret complex scenes more effectively, thereby increasing the precision of instance segmentation and detection.

\textbf{Semantic Segmentation}. 
Much like instance segmentation, semantic segmentation is a pixel-level classification task where image data leads to extracting intricate image features. However, it faces unique challenges, particularly in scenarios where the background closely resembles foreground objects, complicating their differentiation. In multi-task networks, the efficacy of semantic segmentation can be substantially enhanced. This improvement is achieved by exploiting the synergies across different tasks and implementing task-specific optimization strategies, such as fine-tuning the loss function. These approaches collectively bolster the accuracy and robustness of semantic segmentation, enabling more precise and reliable classification in complex visual environments.

\textbf{Lane Segmentation}. 
Lane segmentation is a distinct task in networks like YOLOP, HybridNets, and YOLOPv2. It requires nuanced feature details, which is why these networks source input for the lane segmentation head from deeper layers of the neck. HybridNets uses Tversky loss with Focal loss, while YOLOPv2 uses Dice loss with Focal loss to address challenges related to hard examples and voxel imbalance. In contrast, networks like BEVerse and BEVFormer generate BEV features for segmentation tasks and incorporate lanes as a specific segmentation category, allowing for indirect lane identification.

\textbf{Drivable Area Segmentation}. 
Similar to lane segmentation, drivable area segmentation in models like YOLOP, HybridNets, and YOLOPv2 is approached as a separate task. This segmentation also requires detailed feature input. Hence in YOLOP, the drivable area segmentation head derives its input from the final layer of the FPN. However, with this approach, YOLOPv2 observed negligible performance gains and increased computational cost. As a result, the input for the drivable area head was shifted to an earlier stage in the FPN, complemented by additional upsampling layers in the backbone to mitigate potential feature loss. Networks that perform segmentation via BEV, such as BEVerse, also predict the drivable area as a category, delineating feasible regions in the BEV segmentation output.

\textbf{Depth Estimation}. 
Integrating deep learning techniques with image data has catalyzed remarkable progress in depth estimation. MGnet for example, utilizes the Dense Geometrical Constraints Module (DGC)~\cite{schon2021mgnet} to predict depths in panoramic images, subsequently transforming these estimations into 3D point cloud representations. This method offers a more nuanced understanding of complex environments. Similarly, Ominet adopts fisheye camera images, employing specialized algorithms~\cite{kumar2021omnidet} to leverage the lenses' extensive field of view for depth estimation. Another notable development is the model like LSS~\cite{philion2020lift}, BEVDet~\cite{huang2023detecting}, BEVFusion~\cite{liu2022bevfusion}, and BEVFormer~\cite{li2022bevformer}, which constructs BEV feature maps from multi-view images, enhancing depth estimation accuracy by harnessing multiple perspectives. Such innovations are particularly crucial in fields like autonomous driving.
However, these advancements are not without challenges. A primary limitation is the inherent 2D nature of images, which lack direct 3D information, necessitating sophisticated algorithms for depth inference. Furthermore, the quality and accuracy of image-based depth estimation are often vulnerable to environmental variables, such as fluctuating lighting conditions and changes in viewpoint, leading to potential inconsistencies in perception. Additionally, images' intrinsic resolution and field of view limitations may constrain their effectiveness, especially in situations requiring wide-ranging or highly detailed environmental views.

\subsection{Point Cloud-based Network}
Networks that rely on image input alone are insufficient when it comes to obtaining 3D information. Additionally, the camera's performance is greatly impacted by environmental factors such as lighting and weather. This ultimately affects the accuracy of the model. On the other hand, LiDAR technology has a clear advantage as it provides precise 3D information and demonstrates superior measurement accuracy and robustness. LiDAR technology is capable of providing stable perception performance even in challenging conditions such as nighttime or extreme weather scenarios.
 
\subsubsection{Backbone}
\ 
\newline
Diverse strategies are employed for extracting and processing point cloud data, each offering distinct benefits and facing specific challenges. 

Direct operation on original 3D point clouds is a feature of some deep learning architectures, allowing immediate interaction with complex spatial data. This approach is particularly advantageous for tasks requiring an intricate understanding of 3D structures. Among these methods, 3D convolution techniques, such as the Unet3D~\cite{cciccek20163d} model utilized in LiDARMTL, are designed to extract rich 3D features. However, the primary trade-off of this method lies in its substantial computational demand. On the other hand, Sparse 3D CNNs, as implemented in LiDARMultiNet, present a more resource-efficient solution. These networks are engineered for sparsely populated data, focusing convolution operations solely on non-zero data points, which markedly reduces the computational requirements.
\begin{table}[!htb]
\begin{tabular}{lllcc}
\hline
Input Modality & Model basis        & Techniques   & Backbone              & Neck                 \\ \hline
\multicolumn{1}{c}{LiDAR Point Cloud} & CNN-based & LiDARMTL~\cite{feng2021simple}     & UNet3D~\cite{cciccek20163d}                                                                                 & N/A \\ \cline{3-5} 
                                      &           & AOP-Net~\cite{xu2023aop}      & \begin{tabular}[c]{@{}c@{}}Dual-task 3D Backbone \\ ConvMLP (SC) Backbone\end{tabular} & IRF \\
               &                    &               & \multicolumn{1}{l}{}  & \multicolumn{1}{l}{} \\ \cline{2-5} 
               & CNN \& Transformer & LiDARFormer~\cite{zhou2023lidarformer}   & VoxelNet~\cite{zhou2018voxelnet}              & XSF                  \\ \cline{3-5} 
               &                    & LiDARMultiNet~\cite{ye2023lidarmultinet} & 3D Sparse Convolution~\cite{graham2017submanifold}~\cite{graham2015sparse} & GCP                  \\ \cline{3-5} 
                                      &           & SphereFormer~\cite{lai2023spherical} & \begin{tabular}[c]{@{}c@{}}U-net~\cite{ronneberger2015u}\\ VoxelNet~\cite{zhou2018voxelnet}\\ PointPillars~\cite{lang2019pointpillars}\end{tabular}                & N/A \\ \hline
\end{tabular}
\caption{Backbone and neck for LiDAR-based models}
\label{table:Backbone and neck for LiDAR-based Models}
\end{table}
\vspace{-18pt}

Voxelization strategies, exemplified by VoxelNet and PointPillars, provide another efficient pathway. By dividing point cloud data into structured 3D grids or voxels, these methods allow for applying conventional CNNs in feature extraction, effectively balancing processing efficiency with detail capture. Moreover, some models opt for transforming 3D point clouds into 2D image planes, such as BEV maps. This transformation simplifies the data, making it more amenable to established 2D processing techniques.

The primary challenge for these transformation strategies lies in preserving the rich information content of the original point cloud data. This includes managing the intricacies related to dimensionality, density variations, and the scale of the data post-transformation. Successfully navigating these challenges is crucial for maintaining the fidelity of the 3D information in the transformed domain, ensuring accurate and reliable perception in applications like autonomous driving.
\subsubsection{Neck}
\ 
\newline
Innovative methods have been developed to enhance feature extraction in the realm of neck in panoptic perception systems. AOP-Net~\cite{xu2023aop} introduces the Instance-based Feature Retrieval (IFR) method, aimed at enriching coarse-scale features, thus improving the granularity of the output. LiDARMultiNet~\cite{ye2023lidarmultinet} adopts Global Context Pooling (GCP), functioning similarly to the Global Context Module (GCM), to capture and integrate global contextual information within the network. Additionally, LiDARFormer~\cite{zhou2023lidarformer} utilizes the Cross-space Transformer (XSF), an advanced mechanism for more effective extraction and sharing of global features across different data spaces.
\subsubsection{Head}
\ 
\newline
In this survey, the six tasks under study implement specialized heads designed to complete their respective tasks by leveraging the features extracted and enhanced by the backbone and neck components. This approach differs from the image-based input processing discussed in section~\ref{image-based head}.
\begin{table}[!htb]
\begin{tabular}{lllclcllc}
\hline
Input Modality & Model basis & Techniques & \multicolumn{1}{l}{Tasks} &  & \multicolumn{1}{l}{} &  &  & \multicolumn{1}{l}{} \\ \cline{4-9} 
 &
   &
   &
  \multicolumn{1}{l}{OD} &
  LS &
  \multicolumn{1}{l}{DAS} &
  IS &
  SS &
  \multicolumn{1}{l}{DE} \\ \hline
\multicolumn{1}{c}{LiDAR Point Cloud} &
  CNN-based &
  LiDARMTL &
  \checkmark &
   &
  \checkmark &
   &
   &
  \checkmark \\
 &
   &
  AOP-Net &
  \checkmark &
   &
   &
  \multicolumn{1}{c}{\checkmark} &
  \multicolumn{1}{c}{\checkmark} &
  \checkmark \\
 &
   &
   &
  \multicolumn{1}{l}{} &
   &
  \multicolumn{1}{l}{} &
   &
   &
  \multicolumn{1}{l}{} \\ \cline{2-9} 
 &
  CNN \& Transformer &
  LiDARFormer &
  \checkmark &
   &
  \checkmark &
  \multicolumn{1}{c}{\checkmark} &
  \multicolumn{1}{c}{\checkmark} &
  \checkmark \\
 &
   &
  LiDARMultiNet &
  \checkmark &
   &
  \checkmark &
  \multicolumn{1}{c}{\checkmark} &
  \multicolumn{1}{c}{\checkmark} &
  \checkmark \\
 &
   &
  SphereFormer &
  \checkmark &
   &
  \checkmark &
   &
  \multicolumn{1}{c}{\checkmark} &
  \checkmark \\ \hline
\end{tabular}
\caption{Head for LiDAR-based Models}
\label{table:Head for LiDAR-based Models}
\end{table}
\vspace{-18pt}

\textbf{Object Detection}.
In the domain of 3D object detection, methodologies can be broadly classified into four categories based on their stage and use of anchors, as previously delineated. For instance, PointPillars represents a one-stage, anchor-based approach, efficiently processing point cloud data for object detection. On the other hand, CenterPoint exemplifies a single-stage, anchor-free detection network, offering an alternative strategy for identifying objects within LiDAR point clouds. Two-stage, anchor-based algorithms, such as MV3D and AVOD, provide a more layered approach to detection, involving initial region proposals followed by refined object localization. A key aspect of object detection using LiDAR point clouds is handling the intricacies of point cloud data and accurately predicting 3D bounding boxes. In addition, techniques like Non-Maximum Suppression (NMS) play a pivotal role in the post-processing stage, essential for minimizing redundant and overlapping predictions and enhancing the overall precision of the detection system.

\textbf{Instance Segmentation $\&$ Semantic Segmentation}. Tackling instance segmentation in point clouds presents unique challenges, primarily due to the data’s unordered and sparse nature. Predominant methods include point-based and voxel-based instance segmentation techniques. These approaches typically reconceptualize the task as a point cloud clustering challenge, employing unsupervised learning techniques like Graph Convolutional Networks (GCN)~\cite{lin2020convolution}~\cite{wang2019graph} and spectral clustering. The loss functions in these methods often encompass segmentation loss and center point offset loss, supplemented by loss functions such as mutual information loss to refine clustering effectiveness. Similar to instance segmentation, the primary challenge in semantic segmentation with point cloud data is extracting meaningful features from the inherently unordered structure. The loss function for the segmentation task generally involves multi-class cross-entropy loss, often augmented with regularization terms like Dice loss to enhance the segmentation performance.

\textbf{Lane Segmentation}. Research into lane segmentation using LiDAR point clouds is relatively nascent. Emerging methodologies often involve projecting point clouds onto a BEV plane, followed by lane segmentation on this transformed view. The loss functions for these methods might extend beyond the typical segmentation and detection losses to include BEV-specific segmentation losses, reflecting the unique requirements of lane analysis in point cloud data.

\textbf{Drivable Area Segmentation}. For drivable area segmentation, point cloud data offers valuable depth and shape information, enhancing the task's accuracy. By leveraging the height attributes within the point cloud, ground, and non-ground areas can be differentiated to delineate drivable zones. The primary loss function used here is usually binary cross-entropy loss, which can be further optimized by integrating regularization terms such as Dice loss to improve the segmentation results.

\textbf{Depth Estimation}. 
While point cloud data inherently provides direct depth information for each point, certain scenarios necessitate additional processing or optimization of this depth data. This is particularly relevant when generating depth maps or undertaking point cloud reconstruction tasks. Specialized depth processing techniques are often employed to refine depth estimation in these contexts. For instance, deep learning models can be utilized to discern and learn the intricate relationships between depth and color attributes in the data, thereby enhancing the accuracy and reliability of depth estimations. In such applications, the choice of loss function is pivotal. Typically, a pixel-level squared error loss is used, which effectively measures and minimizes the discrepancies between the predicted depth values and the actual depth measurements.

\subsection{Fusion of Image and Point Cloud}  \label{sensor fusion network}
LiDAR's inability to capture color and texture information poses limitations for tasks requiring detailed visual cues. The fusion of image and LiDAR data in panoptic perception techniques aims to harness the complementary strengths of these two data modalities. The primary challenge in this approach is devising an effective strategy for integrating the rich yet distinct information from both sources. This involves ensuring accurate data alignment and synchronization and managing the complexity inherent in processing and learning from these diverse data types. Achieving a harmonious balance between the detailed texture and color information from images and the precise depth and spatial information from LiDAR is key to maximizing the efficacy of panoptic perception systems.
\subsubsection{Backbone and Neck}
\ 
\newline
As detailed in section~\ref{sensorfusion}, mid-term fusion is currently a prevalent technique in sensor fusion, effectively combining image and LiDAR data at an intermediate stage within the network architecture. This approach is exemplified in networks such as BEVFusion~\cite{liu2022bevfusion} and CALICO~\cite{sun2023calico}, where features from distinct sources are integrated to harness their combined strengths. In these networks, the Swin-Transformer is employed for its adeptness in extracting intricate image features, while VoxelNet and PointPillars are utilized for their proficiency in processing point cloud data, subsequently transforming these into 2D feature maps.

This fusion process predominantly occurs on the BEV feature maps, merging image and point cloud features. Such integration results in a more comprehensive and enriched feature set, encapsulating a wider range of environmental cues and details. BEVFusion further augments this process by incorporating a Feature Pyramid Network (FPN) to enhance image-derived features, ensuring a more robust and detailed representation.
\begin{table}[!htb]
\begin{tabular}{lllcc}
\hline
Input Modality & Model basis & Techniques    & Backbone             & Neck                 \\ \hline
Sensor Fusion  & CNN \& Transformer & BEVFusion~\cite{liu2022bevfusion} & \begin{tabular}[c]{@{}c@{}}Swin-transformer\\ VoxelNet\end{tabular}     & FPN \\ \cline{3-5} 
 &                    & CALICO~\cite{sun2023calico}    & \begin{tabular}[c]{@{}c@{}}Swin-transformer\\ PointPillars\end{tabular} & N/A \\ \hline
\end{tabular}
\caption{Backbone and neck for multi-modal models}
\label{table:Backbone and Neck for Multimodel-based Models}
\end{table}
\vspace{-20pt}

Fusing these features in a mid-term stage allows for a more nuanced understanding of the environment by leveraging the depth and spatial accuracy of LiDAR with the textural and color detail provided by images. This synergy is crucial in applications like autonomous driving, where a detailed and accurate perception of the surroundings is paramount. However, the challenge lies in effectively aligning and integrating these diverse data types, requiring sophisticated methodologies to ensure seamless fusion without information loss.
\subsubsection{Head}
\ 
\newline
\textbf{Object Detection}. 
Object detection leveraging images and LiDAR data capitalizes on the synergy of LiDAR's precise 3D spatial information and the rich textural detail that images offer. In multi-modal object detection models, the focus is typically on 3D object detection, where methods discussed earlier remain applicable and effective. For instance, Transfusion~\cite{bai2022transfusion} adopts a transformer-based approach for 3D object detection. BEVDet~\cite{huang2023detecting} employs a two-stage, anchor-free methodology, initially generating proposals followed by extracting corresponding heatmaps for accurate localization of the objects.
In the MVP framework~\cite{yin2021multimodal}, image feature extraction utilizes CenterNet's capabilities, while LiDAR features are processed through either VoxelNet or PointPillars. This process culminates in mapping 2D image feature points onto sparse point cloud features, facilitating a comprehensive fusion for object detection. The loss functions commonly employed in these tasks include cross-entropy loss for object classification and smooth L1 loss for bounding box regression, ensuring precise object localization and categorization within the multimodal detection framework.
\begin{table}[!htb]
\begin{tabular}{lllllllll}
\hline
Input Modality & Model basis & Techniques & Tasks &       &     &    &    &    \\ \cline{4-9} 
               &             &  & OD    & LS & DAS & IS & SS & DE \\ \hline
Sensor Fusion & CNN \& Transformer & BEVFusion & \multicolumn{1}{c}{\checkmark} & \multicolumn{1}{c}{\checkmark} & \multicolumn{1}{c}{\checkmark} &  & \multicolumn{1}{c}{\checkmark} & \multicolumn{1}{c}{\checkmark} \\
              &                    & CALICO    & \multicolumn{1}{c}{\checkmark} &                       & \multicolumn{1}{c}{\checkmark} &  & \multicolumn{1}{c}{\checkmark} & \multicolumn{1}{c}{\checkmark} \\ \hline
\end{tabular}
\caption{Head for Multimodal-based Models}
\label{table:Head for Multimodel-based Models}
\end{table}
\vspace{-22pt}

\textbf{Instance Segmentation $\&$ Semantic Segmentation}. The fusion of the two types of sensors can help the system better distinguish objects and backgrounds, clarify the boundaries of objects, and understand the structure of the scene in 3D space. When using image and LiDAR for fusion to handle instance segmentation tasks, one strategy is that the depth information of LiDAR can be combined with the high-resolution visual details of the image to distinguish objects close to each other through 3D spatial clustering and project them to 2D images to obtain more accurate instance boundaries. For example, two closely parked cars may be difficult to separate in an RGB image, but with the depth data provided by LiDAR, they can be separated in 3D space and accurately segmented on the image~\cite{geng2020deep}. Another strategy is to design a two-stream network~\cite{kim2018season}~\cite{caltagirone2018lidarcamera}, with one stream processing image data and the other stream processing LiDAR data, and then at some stage, the two streams are fused for decision-making. In this way, each data type can use the feature extraction method most suitable for its characteristics. For example, the BEV map is to convert two different types of features into BEV map for fusion and then perform specific segmentation tasks. Therefore, the fusion approach combines the strengths of both sensors, allowing the model to rely on depth information to improve segmentation accuracy in regions with similar textures or blurred colors.

\textbf{Lane Segmentation}. 
Fusing image and LiDAR data for lane segmentation can greatly improve the accuracy of detecting lanes. This is because the combination of rich texture and color information from images and precise depth information from LiDAR creates a powerful synergy. One approach is to project LiDAR data onto the image plane to create a depth image that is then merged with the RGB image for more detailed lane segmentation. This technique has been shown to be effective in prior studies~\cite{hu2014multi}~\cite{caltagirone2018lidarcamera}. Alternatively, lane segmentation can be performed on BEV representations, which provide a clearer overall view of lanes. The lane segmentation process in this approach is similar to traditional segmentation techniques. However, processing on the BEV map may require specific computational losses associated with this representation.

\textbf{Drivable Area Segmentation}. 
For the drivable area segmentation task, the LiDAR data can be projected onto the image plane to generate a depth map, and then the depth map and the RGB image are fused to form a new input, which is used to segment the drivable area. This strategy can combine the detailed texture information of the image and the accurate depth information of LiDAR to identify the drivable area better. At the same time, the drivable area segmentation can also be carried out in the BEV representation because the BEV representation can also show the whole picture of the drivable area.

\textbf{Depth Estimation}. While LiDAR data inherently provides precise depth information, image data can be employed to refine this depth estimation further. By utilizing the texture and color information captured in images, it is possible to correct and enhance the depth details from LiDAR data. Concurrently, LiDAR data can serve as a reliable ground truth in training depth estimation models, ensuring their accuracy and robustness. This synergistic use of both data types allows for more nuanced depth perception, leveraging the strengths of each method.

\section{Comparison With Other Perception Techniques} \label{Comparison With Other Perception Techniques}

\begin{table}[!htb]
\centering
\begin{tabular}{ccccccc}
\hline
Model        & FPS  & Params(M) & GFLOPs & AP50(\%) & mIoU-d(\%) & IoU-l(\%) \\ \hline
YOLOv5s\cite{yolov5s}      & 82.0   & 7.2       & 16.5   & 77.2     & -          & -         \\
Sparse RCNN\cite{sun2021sparse}  & 20.0   & 77.8      & 23.3  & 81.9     & -          & -         \\
Enet\cite{paszke2016enet}         & 100.0  & -         & -      & -        & -          & 14.6     \\
ENet-SAD\cite{hou2019learning}     & 50.6 & 1.0      & -      & -        & -          & 16.0     \\
GCNet\cite{cao2019gcnet}        & 30.1 & 28.1     & -      & -        & 82.1      & -         \\
DNLNet\cite{yin2020disentangled}       & 28.6 & 71.5         & 765.2      & -        & 84.4      & -         \\
PSPNet\cite{zhao2017pyramid}       & 11.1 & -         & -      & -        & 89.6       & -         \\ \midrule
DeepLabV3+\cite{chen2018encoderdecoder}   & 23.4 & 15.4      & 30.7   & -        & 90.9       & 29.8      \\
SegFormer\cite{xie2021segformer}    & 30.8 & 7.2       & 12.1   & -        & 92.3       & 31.7      \\
DLT-Net\cite{qian2019dlt}      & 9.3  & -         & -      & 68.4     & 71.3       & -         \\
MultiNet\cite{teichmann2018multinet}     & 8.6  & -         & -      & 60.2     & 71.6       & -         \\
YOLOP\cite{wu2022yolop}        & 24.0   & 7.9       & 18.6   & 76.5     & 91.5       & 26.2      \\
HybridNets\cite{vu2022hybridnets}   & 26.0   & 12.8     & 15.6   & 77.3     & 90.5       & 31.6      \\
YOLOPv2\cite{han2022yolopv2}      & -    & 38.9      & -      & 83.4     & 93.2       & 27.3     \\
CenterPNets\cite{chen2023centerpnets}  & -    & 28.6         & -      & 81.6     & 92.8       & 32.1      \\
Sparse U-PDP\cite{wang2023sparse} & 29.0   & 12.1     & 15.1   & 84.1     & 92.9       & 32.0      \\ \hline
\end{tabular}
\caption{Multiple task comparisons of different models on BDD100K}
\label{table:comparision image based}
\end{table}
\vspace{-18pt}
This section presents a detailed comparative analysis of SOTA models in the domains of multi-task learning and panoptic perception, scrutinizing them across two critical dimensions: performance and efficiency. For the evaluation of image-based panoptic perception models, BDD100K dataset is used. The FPS results displayed in table~\ref{table:comparision image based} are derived from the study by Wang et al.~\cite{wang2023sparse}, while other evaluation metrics are sourced directly from the respective models. In examining models that rely on LiDAR and multi-modal inputs, the nuScenes dataset serves as the foundational dataset for our analysis, given its widespread adoption in multi-modal and multi-task model evaluations. The results for LiDAR-based and multi-modal models, as shown in table~\ref{table:Comparisons of Different Models on nuScenes}, are based on the findings from Liu et al.\cite{liu2022bevfusion} and the original publications.

In addition to the panoptic perception models discussed in section~\ref{Techniques for Panoptic Perception}, these tables also select typical single-task models that are pertinent to specific perception tasks. These models are included to facilitate a more direct and intuitive comparison between the capabilities of multi-task and single-task models in perception-related tasks. This comparative approach aims to provide a comprehensive perspective on the current state and effectiveness of various model architectures in the field of autonomous driving perception.
\begin{table}[!htb]
\centering
\begin{tabular}{ccccccc}
\hline
Model         & Modality & FPS   & Params(M) & GFLOPs & mAP   & NDS  \\ \hline
SECOND\cite{yan2018second}        & L        & 14.3  & -          & 170.0       & 0.528 & 0.633            \\
TransFusion\cite{bai2022transfusion}   & C+L      & 6.39  & -          & 971.6       & 0.689 & 0.716            \\ \midrule
PETRv2\cite{liu2022petrv2}        & C        & -      & -          & -       & 0.490 & 0.582     \\
BEVFormer\cite{li2022bevformer}     & C        & -      & 68.7      & 1303.5 & 0.412 & 0.520    \\
BEVFormerv2\cite{yang2023bevformer}   & C        & -      & -          & -       & 0.556 & 0.634            \\
$M^{2}BEV$\cite{xie2022m}         & C        & -      & >112.5           & -       & 0.425 & 0.465        \\
AOP-Net\cite{xu2023aop}       & L        & -      & 14.6      & -       & 0.582 & 0.657            \\
LiDARFormer\cite{zhou2023lidarformer}   & L        & -      & 77.0      & -       & 0.715 & 0.743        \\
LiDARMultiNet\cite{ye2023lidarmultinet} & L        & -      & 131.0     & -       & 0.670 & 0.716        \\
SphereFormer\cite{lai2023spherical}  & L        & -      & -          & -       & 0.685 & 0.728        \\
BEVFusion\cite{liu2022bevfusion}     & C+L      & 8.4   & -          & 506.4       & 0.702 & 0.729       \\
CALICO\cite{sun2023calico}        & C+L      & -      & -          & -       & 0.627 & 0.601       \\ \hline
\end{tabular}
\caption{Detection comparisons of different models on nuScenes.}
\label{table:Comparisons of Different Models on nuScenes}
\end{table}
\vspace{-20pt}

\subsection{Performance Analysis}
Performance evaluation in autonomous driving models is typically delineated by specific indicators corresponding to distinct tasks. The primary metrics of AP50 and mAP are employed for object detection, while the nuScenes dataset introduces the comprehensive NDS indicator for a more holistic model assessment. Segmentation tasks, such as lane segmentation, drivable area segmentation, and BEV segmentation, are quantitatively measured using the mIoU metric.

As demonstrated in table~\ref{table:comparision image based} and table~\ref{table:Comparisons of Different Models on nuScenes}, it is observed that specific multi-task models outperform their single-task models. Notably, YOLOPv2 and Sparse U-PDP excel in three tasks, surpassing the performance of single-task models on BDD100K. In the nuScenes dataset, LiDARFormer and BEVFusion exceed single-task models on the 3D object detection task regarding both mAP and NDS.

In the evaluation of BEV segmentation tasks within the nuScenes dataset, it is noteworthy that not all models test the full range of categories. To ensure a comprehensive comparison, specific category results are detailed in table~\ref{table:segmentation compare}. As per the data from ~\cite{liu2022bevfusion} and ~\cite{sun2023calico}, both BEVFusion and Calico demonstrate superior performance over single-task segmentation models, namely OFT~\cite{roddick2018orthographic}and CVT~\cite{zhou2022crossview}. This enhanced performance can be attributed to the inherent capability of multi-task models to share underlying feature representations. This sharing enables the models to leverage valuable insights across tasks, enriching the overall segmentation process. For instance, common low-level features can significantly benefit both aspects in combined object detection and segmentation tasks.
\begin{table}[htb]
\centering
\resizebox{\columnwidth}{!}{%
\begin{tabular}{@{}lcccccccc|cc@{}}
\toprule
\multicolumn{1}{c}{Model} &
  \multicolumn{1}{l}{PETRv2} &
  \multicolumn{1}{l}{BEVFormer} &
  \multicolumn{1}{l}{$M^{2}BEV$} &
  \multicolumn{1}{l}{LiDARFormer} &
  \multicolumn{1}{l}{LiDARMultiNet} &
  \multicolumn{1}{l}{SphereFormer} &
  \multicolumn{1}{l}{CALICO} &
  \multicolumn{1}{l|}{BEVFusion} &
  \multicolumn{1}{l}{OFT} &
  \multicolumn{1}{l}{CVT} \\ \midrule
\multicolumn{1}{c}{mIoU(\%)}                                                             & 60.3 & 48.7 & 57.0 & 81.0 & 81.4 & 81.9 & 56.7 & 62.7 & 42.1 & 40.2 \\ \midrule
\multicolumn{1}{l|}{barrier}                                                         & -    & -    & -    & 83.5 & 80.4 & 83.3 & -    & -    & -    & -    \\
\multicolumn{1}{l|}{bicycle}                                                         & -    & -    & -    & 39.8 & 48.4 & 39.2 & -    & -    & -    & -    \\
\multicolumn{1}{l|}{bus}                                                             & -    & -    & -    & 85.7 & 94.3 & 94.7 & -    & -    & -    & -    \\
\multicolumn{1}{l|}{car}                                                             & -    & 46.8 & -    & 92.4 & 90.0 & 92.5 & -    & -    & -    & -    \\
\multicolumn{1}{l|}{\begin{tabular}[c]{@{}l@{}}construction \\ vehicle\end{tabular}} & 46.3 & 46.7 & -    & 70.8 & 71.5 & 77.5 & -    & -    & -    & -    \\
\multicolumn{1}{l|}{motorcycle}                                                      & -    & -    & -    & 91.0 & 87.2 & 84.2 & -    & -    & -    & -    \\
\multicolumn{1}{l|}{pedestrian}                                                      & -    & -    & -    & 84.0 & 85.2 & 84.4 & -    & -    & -    & -    \\
\multicolumn{1}{l|}{traffic cone}                                                    & -    & -    & -    & 80.7 & 80.4 & 79.1 & -    & -    & -    & -    \\
\multicolumn{1}{l|}{trailer}                                                         & -    & -    & -    & 88.6 & 86.9 & 88.4 & -    & -    & -    & -    \\
\multicolumn{1}{l|}{truck}                                                           & -    & -    & -    & 73.7 & 74.8 & 78.3 & -    & -    & -    & -    \\
\multicolumn{1}{l|}{\begin{tabular}[c]{@{}l@{}}driveable \\ surface\end{tabular}}    & 85.6 & 77.5 & 75.9 & 97.8 & 97.8 & 97.9 & 82.4 & 85.5 & 74.0 & 74.3 \\
\multicolumn{1}{l|}{other flat}                                                      & -    & -    & -    & 69.0 & 67.3 & 69.0 & -    & -    & -    & -    \\
\multicolumn{1}{l|}{side walk}                                                       & -    & -    & -    & 80.9 & 80.7 & 81.5 & 63.1 & 67.6 & 45.9 & 39.9 \\
\multicolumn{1}{l|}{terrain}                                                         & -    & -    & -    & 76.9 & 76.5 & 77.2 & -    & -    & -    & -    \\
\multicolumn{1}{l|}{manmade}                                                         & -    & -    & -    & 91.9 & 92.1 & 93.4 & -    & -    & -    & -    \\
\multicolumn{1}{l|}{vegetation}                                                      & -    & -    & -    & 89.0 & 89.6 & 90.2 & -    & -    & -    & -    \\
\multicolumn{1}{l|}{lane}                                                            & 49.0 & 23.9 & 38.0 & -    & -    & -    & 44.9 & 53.7 & 33.9 & 29.4 \\
\multicolumn{1}{l|}{ped. cross.}                                                     & -    & -    & -    & -    & -    & -    & 57.0 & 60.5 & 35.3 & 36.8 \\
\multicolumn{1}{l|}{stop line}                                                       & -    & -    & -    & -    & -    & -    & 41.3 & 52.0 & 27.5 & 25.8 \\
\multicolumn{1}{l|}{carpark}                                                         & -    & -    & -    & -    & -    & -    & 51.6 & 57.0 & 35.9 & 35.0 \\ \bottomrule
\end{tabular}%
}
\caption{Segmentation comparisons of different models on nuScenes.}
\label{table:segmentation compare}
\end{table}
\vspace{-15pt}

Moreover, BEVFusion and Calico, as multi-modal models, gain an additional advantage from incorporating LiDAR data, a feature absent in OFT and CVT. This integration results in more precise segmentation outcomes. The superiority of LiDAR, particularly in segmenting specific types like drivable surfaces and sidewalks, is evident in the comparison presented in table~\ref{table:segmentation compare}. 

\subsection{Efficiency Analysis}
This study evaluates model efficiency by analyzing responsiveness, measured in latency and FPS, and resource utilization, assessed through parameter size and GFLOPs. Collectively, these metrics provide insights into the model's inference speed and computational resource demands.

\subsubsection{Responsiveness}
\begin{figure}[htbp]
  \centering
      \subfigure[Latency comparison] {\includegraphics[width=.48\textwidth]{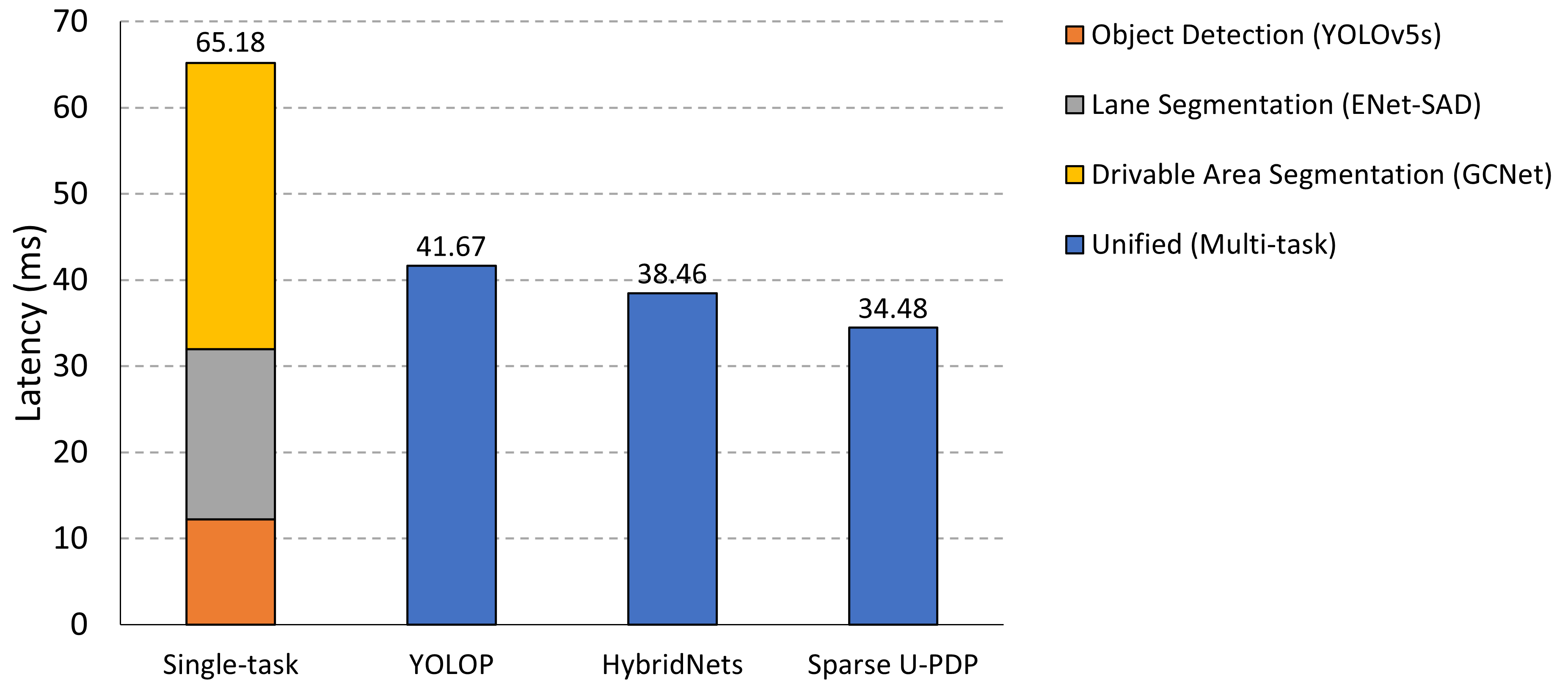}\label{fig:task compare}}
      \hspace{2mm}
      \subfigure[Parameter size comparison]{\includegraphics[width=.48\textwidth]{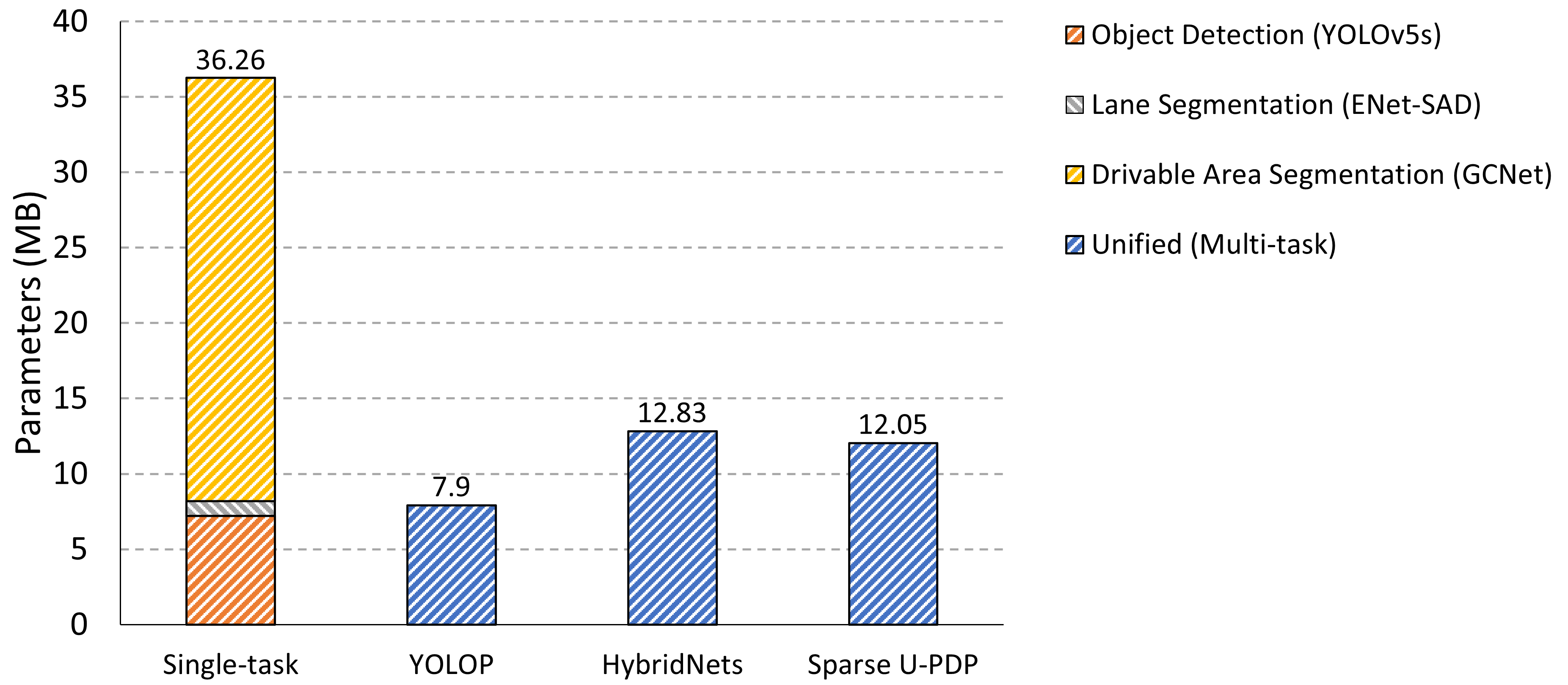}\label{fig:parameter compare}}
  \caption{Comparison between single-task combination and multi-task models.}
  \label{fig:comparison}
\end{figure}

\ 
\newline
The analysis of model responsiveness is primarily measured using Frames Per Second (FPS). To ensure a fair comparison across different models, testing them on identical hardware setups is compulsory. According to ~\cite{wang2023sparse}, the FPS data presented in table~\ref{table:comparision image based} were obtained using a single Nvidia RTX 3090. Similarly, the FPS results for models in table~\ref{table:Comparisons of Different Models on nuScenes}, as reported by ~\cite{liu2022bevfusion}, were also evaluated on the Nvidia RTX 3090 hardware.
An observation from table~\ref{table:comparision image based} reveals that single-task models generally exhibit higher FPS. However, it is crucial to recognize that multi-task models concurrently deliver outputs for multiple tasks, making a direct FPS comparison somewhat unfair. For a more unbiased assessment of real-time performance, we adopt a straightforward approach by summing the latencies of different single tasks to approximate the total latency or FPS of a composite model. This methodology is grounded in that autonomous driving systems comprising single tasks are typically modular.
In this analysis, the most efficient single-task models from table~\ref{table:comparision image based} are combined to form a unified perception model, encompassing YOLOv5s, ENet-SAD, and GCNet. The aggregate latency of this combined model approximates to 65ms, correlating to an effective FPS of about 18. As illustrated in Figure~\ref{fig:task compare}, the latency of multi-task models such as YOLOP~\cite{wu2022yolop}, HybridNets~\cite{vu2022hybridnets}, and Sparse U-PDP~\cite{wang2023sparse} is observed to be less than 65ms, and their FPS greater than 18, thereby underscoring the enhanced real-time performance of multi-task models.


The models in table~\ref{table:Comparisons of Different Models on nuScenes} have a low FPS because they perform 3D detection tasks. On the other hand, the models in table~\ref{table:comparision image based} are 2D based and thus have a higher FPS. The 3D detection task is more complex as it needs to take into account several factors such as occlusion, object diversity, rotation, scale changes, etc. This complexity increases the number of parameters and calculation requirements. Additionally, the data involved in 3D object detection, such as LiDAR point clouds, is usually huge. Each frame may contain tens of thousands to millions of points, and each point may have multiple attributes such as location, reflection intensity, etc. The LiDAR point cloud data needs to go through several pre-processing steps, further increasing the processing time.
\vspace{-18pt}

\subsubsection{Resource Utilization}
\ 
\newline
Resource occupation analysis can be determined by the size of parameters and GFLOPs. In autonomous driving systems, besides solving perception tasks, tasks such as motion planning and localization also require a significant amount of computing resources. Therefore, if a model is too complex, it may consume more resources, which can adversely affect other tasks. The parameter size of the models in figure~\ref{fig:task compare} is evaluated, and the results are shown in figure~\ref{fig:parameter compare}. Similarly, for a fair comparison, the parameter sizes of the single-task model are also accumulated and compared with the parameter sizes of the multi-task model. YOLOP~\cite{wu2022yolop}, HybridNets~\cite{vu2022hybridnets}, and Sparse U-PDP~\cite{wang2023sparse} have smaller parameter sizes as multi-task models, which is expected because all heads in multi-task models share a backbone. This further demonstrates that a multi-task model can significantly reduce unnecessary computing resources.
Furthermore, the GFLOPs metric, as presented in table~\ref{table:comparision image based}, is a crucial indicator for evaluating the computational complexity of various models. This table shows that the GFLOPs for multi-task models like SegFormer~\cite{xie2021segformer}, HybridNets~\cite{vu2022hybridnets}, and Sparse U-PDP~\cite{wang2023sparse} are lower than the single-task detection model YOLOv5s~\cite{yolov5s}. This indicates a remarkable level of computational efficiency in these multi-task models.
In contrast, the models listed in table~\ref{table:Comparisons of Different Models on nuScenes} generally exhibit higher parameter sizes and GFLOPs. This can be primarily attributed to their more complex calculations, such as 3D object detection and the construction of BEV feature maps. Consequently, there is an evident need for further optimization of these models or algorithms to align them with the requisite performance standards regarding computational efficiency and resource utilization.
\section{Challenges and Future Directions} \label{Challenges and Future Directions}
The development of multi-task perception networks for autonomous driving is facing many challenges, including weight balance, task relevance, and the avoidance of negative transfer. In this section, we introduce those challenges and present solutions for each, as well as propose potential research directions for practitioners.

\subsection{Discussion of challenges and limitations of current techniques}

\textbf{Weight balance} is a crucial issue in multi-task learning. Each task requires different levels of attention and resources, and the weights need to be distributed appropriately to ensure that important tasks receive the necessary focus without neglecting those that are relatively easy but still necessary. This requires evaluating the difficulty and importance of tasks and adjusting the weights dynamically in response to changes in the environment and time.

\textbf{Task relevance} requires a meticulous determination of the dependencies between tasks. An understanding of the dependencies between tasks enables us to design network structures that can effectively coordinate all tasks. However, this process presents a challenge, as it requires an in-depth comprehension of the intricate relationships that may exist between different tasks. It is essential to consider how this understanding can be fed back into the model design and optimization process. This will ensure the development of efficient and effective network structures capable of achieving optimal results.

\textbf{Negative transfer} can be a challenge when optimizing machine learning models. Sometimes, improving the performance of one task can lead to a decrease in performance of other tasks. For instance, if we are optimizing a model for both pedestrian detection and landmark detection, focusing too much on pedestrian detection may cause the model to perform poorly on landmark detection. To prevent negative transfer, it's vital to ensure that the model can balance tasks and maintain its focus on all the tasks at hand.

It is important to consider practical issues such as data bias, model interpretability, training complexity, and computational resource constraints when designing a multi-task perception network for autonomous driving. For example, models should handle imbalanced data and also provide some level of explainability to understand how it works and make improvements. The input source of the network differs significantly in size, structure, and information content between camera data, which provides color and texture information, and LiDAR, which provides precise depth information in the form of point cloud data. Therefore, effectively fusing these two data types is a challenge. In addition, multi-task learning models are often more complex than single-task models, so finding effective ways to optimize the training process is necessary. Under limited computing resources, trade-offs may need to be made to find the best model structure and parameter settings. 

In summary, designing and optimizing a multi-task network for autonomous driving is a complex problem that requires consideration of various factors, including weight balance, task relevance, negative transfer, and other practical problems. Thus, it is necessary to use various techniques and strategies to find the optimal solution.
\vspace{-13pt}
\subsection{Proposed solutions and future research directions}

\textbf{Weight Balancing.} Different methods can be used to better balance the weights of various tasks in panoptic perception models. Traditional methods like grid search or genetic algorithms are time-consuming and require many computing resources. In recent years, gradient-based and uncertainty-based methods have become popular. Uncertainty~\cite{kendall2018multi}, for example, is a multi-objective optimization strategy that balances the importance of different tasks by introducing the uncertainty of each task as a learnable parameter. Through continuous iteration and learning, the uncertainty of each task is adjusted to an appropriate level, and the performance of all tasks is improved in a balanced manner. Another strategy is GradNorm~\cite{chen2018gradnorm}, an optimization strategy for multi-task learning. The main goal of this strategy is to automatically adjust the loss weights for each task to achieve a balance between tasks during training. GradNorm adjusts the loss weight by comparing the size of the gradient norm of different tasks. If the gradient norm of a specific task is more significant than that of other tasks, the weight of this task in the entire optimization process will be reduced, and vice versa.
Weight adjustment strategies still have room for improvement. Future work can explore dynamic weight adjustment strategies~\cite{guo2018dynamic}, \cite{liu2023famo}, \cite{verboven2023hydalearn} or using reinforcement learning with a reward mechanism to learn the weights between tasks automatically.

\textbf{Task relevance.} To address the complexities inherent in task relevance, some researchers have turned to the multi-objective gradient descent algorithm (MGDA), as evidenced by studies like \cite{milojkovic2019multi} and \cite{zhou2021multiple}. MGDA, an approach to multi-objective gradient optimization, is specifically designed to navigate conflicts between various objectives. Diverging from traditional methods that optimize each objective in isolation, MGDA seeks a harmonized solution that aligns with the collective gradient descent direction of all objectives. This methodology facilitates a balanced resolution that not only upholds the performance of individual targets but also mitigates potential conflicts among them.
However, the current scope of task relevance analysis may need to sufficiently encapsulate the intricate relationships among tasks, particularly in the dynamic context of autonomous driving. For instance, the correlation between pedestrian and vehicle detection tasks may vary significantly between urban and rural settings. Thus, future research could focus on devising more sophisticated measures of task correlation. This could include exploring temporal and spatial correlations among tasks and developing adaptive mechanisms to dynamically alter these correlations in response to real-time environmental and situational changes.

\textbf{Negative transfer.} Negative transfer usually occurs when optimizing one task negatively affects other tasks. In the study~\cite{lee2019learning}, researchers introduced a Bayesian-based meta-learning method adept at balancing learning processes across varied tasks, thereby alleviating negative transfer. Further advancing this field, $M^3ViT$ ~\cite{fan2022m3vit} integrates mixture-of-experts (MoE) layers into a vision transformer (ViT) backbone. This design sparsely activates task-specific experts during training, optimizing efficiency and mitigating gradient conflicts between tasks. In inference, the model selectively activates only the relevant sparse expert pathway, streamlining the process and enhancing overall task optimization.
Another promising approach is prompt-learning, as showcased by VE-Prompt ~\cite{liang2023visual}. This method employs visual exemplars as task-specific prompts, guiding the model towards developing high-quality, task-specific representations, effectively reducing the impact of negative transfer and bolstering model performance.
Addressing negative transfer in multi-task learning will require developing sophisticated methods for its detection and quantification. This challenge may necessitate the creation of novel performance metrics and evaluation methodologies. Once negative transfer is reliably identified, the focus can shift to devising strategies to prevent or mitigate it. Potential solutions include refining model architecture, enhancing weight balancing strategies, and optimizing data selection and processing approaches.

\textbf{Other Challenges.} Addressing data bias in multi-task models, familiar to most neural network frameworks, can be approached through established methods. Data augmentation techniques, including random rotation and shearing, offer one viable solution. Additionally, resampling methods like Synthetic Minority Oversampling Technique (SMOTE)~\cite{chawla2002smote} effectively counteract class imbalances. Exploring the potential of deep learning networks to refine data features and mitigate noise also presents a promising direction.
The interpretability of models is another crucial aspect. Tools like TensorBoard~\cite{TensorBoard} and Torchviz~\cite{PyTorchViz} help visualize model processes. Further, methodologies like Local Interpretable Model-Agnostic Explanations (LIME)~\cite{lime} have been proposed to enhance model transparency. It is important to understand the causality and distinct roles of model parameters to design models that are accessible not only to professionals but also to general users. Improving the interpretability of models continues to be a promising field for further research.
To reduce training complexity, optimization algorithms like Adam~\cite{kingma2017adam} and RMSprop~\cite{RMSProp} can expedite the training process. Designing efficient model architectures also can significantly reduce complexity. Additionally, various fusion algorithms are under exploration to integrate camera and LiDAR data effectively. A prevalent method involves leveraging deep learning for feature extraction and fusion, illustrated by multi-modal fusion networks such as BEVFusion~\cite{liu2022bevfusion}, which integrate diverse features in BEV and execute tasks through specific heads. Beyond mid-term feature fusion, early data fusion and late decision fusion are viable strategies in autonomous driving perception tasks.
As large language models gain traction, multi-task models may also expand in size to achieve enhanced performance but are constrained by computational resources. Addressing these limitations, model compression techniques~\cite{han2015deep}~\cite{han2016eie} like pruning~\cite{zhu2017prune}, quantization~\cite{deng2020model}~\cite{choudhary2020comprehensive}, and distillation~\cite{gou2021knowledge} offer potential solutions. Additionally, utilizing distributed computing resources, such as GPU clusters, is vital for efficient model training. During the model's inference phase, strategies involving edge computing or synergy of edge and cloud computing present important research avenues to optimize workload management.
\section{Conclusion} \label{Conclusion}

With the rapid development of autonomous driving technology, panoptic perception has become a hotspot in the research field, providing an all-round perspective for truly unmanned driving. This paper delves into multi-task perception networks in autonomous driving, examining them across diverse modal dimensions, including image-based, point cloud-based, and multi-modal fusion. 
We dissect typical multi-task perception networks, analyzing their architecture from the backbone, neck, and head components. Our examination reveals that each network possesses distinct strengths and limitations in feature extraction, feature enhancement, and executing specific tasks such as object detection, lane segmentation, drivable area segmentation, instance segmentation, semantic segmentation, and depth estimation. Subsequently, a comparative analysis of several key performance and efficiency parameters between multi-task and single-task networks is presented. The findings indicate that SOTA panoptic perception networks not only maintain task accuracy but also excel in reducing latency and optimizing resource utilization. However, the emergence of these networks has also brought to light challenges like weight balancing, task correlation, and negative transfer. Addressing these issues, we collate various effective strategies and propose insightful future research directions.


In general, the domain of panoptic perception in autonomous driving presents a vast landscape filled with both challenges and opportunities. We anticipate that continued research and technological advancements will soon pave the way for more intelligent, safe, and reliable autonomous driving systems, ultimately enhancing the human travel experience.

\section*{Acknowledgment}
This work was partly supported by the U.S. National Science Foundation
under Grants CNS-2245729 and Michigan Space Grant Consortium 80NSSC20M0124.
\bibliographystyle{acm}
\bibliography{reference}
\end{document}